\documentclass[lettersize,journal]{IEEEtran}
 
\usepackage{amsmath,amsfonts}
\usepackage{algorithmic}
\usepackage{algorithm}
\usepackage{graphicx}
\usepackage{cite}

\usepackage[caption=false,font=normalsize,labelfont=sf,textfont=sf]{subfig}
\usepackage{textcomp}
\usepackage{stfloats}
\usepackage{url}
\usepackage{verbatim}

\usepackage{amsfonts}       
\usepackage{nicefrac}       
\usepackage{microtype}      
\usepackage{xcolor}      
\usepackage{threeparttable}
\usepackage{booktabs}
\usepackage{array}
\usepackage{multirow}
\usepackage{arydshln}
\begin{document}

\title{FedMMKT:Co-Enhancing a Server Text-to-Image Model and Client Task Models in Multi-Modal Federated Learning}
\author{
Ningxin He$^{\dagger}$, Yang Liu$^{*}$, Wei Sun, Xiaozhou Ye, Ye Ouyang,~\IEEEmembership{Fellow,~IEEE}, Tiegang Gao$^{*}$, and Zehui Zhang%
\thanks{$^{*}$Corresponding author.}%
\thanks{$^{\dagger}$This research was completed while the first author was an intern at Institute for AI Industry Research, Tsinghua University.}%
\thanks{Ningxin He and Tiegang Gao are with the School of Software Engineering, Nankai University, Tianjin, China (e-mail: 1120220295@mail.nankai.edu.cn; gaotiegang@nankai.edu.cn).}%
\thanks{Yang Liu is with the Department of Computing, Hong Kong Polytechnic University, Hong Kong, China (e-mail: yang-veronica.liu@polyu.edu.hk).}%
\thanks{Wei Sun is with the Institute for AI Industry Research, Tsinghua University, Beijing, China (e-mail: sunwei@air.tsinghua.edu.cn).}%
\thanks{Xiaozhou Ye and Ye Ouyang are with AsiaInfo Technologies, Beijing, China (e-mail: yexz@asiainfo.com; ye.ouyang@asiainfo.com).}%
\thanks{Zehui Zhang is with the China-Austria Belt and Road Joint Laboratory on Artificial Intelligence and Advanced Manufacturing, Hangzhou Dianzi University, Hangzhou, China (e-mail: zhangtianxia918@163.com).}%
}

\maketitle

\begin{abstract}
Text-to-Image (T2I) models have demonstrated their versatility in a wide range of applications. However, adaptation of T2I models to specialized tasks is often limited by the availability of task-specific data due to privacy concerns. On the other hand, harnessing the power of rich multimodal data from modern mobile systems and IoT infrastructures presents a great opportunity. This paper introduces Federated Multi-modal Knowledge Transfer (FedMMKT), a novel framework that enables co-enhancement of a server T2I model and client task-specific models using decentralized multimodal data without compromising data privacy. 
\end{abstract}


\section{Introduction}
\IEEEPARstart{T}{ext} -to-Image (T2I) models such as GLIDE \cite{1}, DALL-E-2 \cite{2}, and Stable Diffusion \cite{3} have seen rapid development across various application domains. To improve the performance of pre-trained T2I models on  specialized domains,targeted fine-tuning is required, but is often limited by privacy constrains of domain-specific data \cite{4}.
Therefore, it is urgent to develop innovative methods for fine-tuning T2I models while ensuring data privacy and security.

Federated Learning (FL) is a paradigm enabling multiple clients to collaboratively train models without directly exchanging data \cite{5,6}. In traditional approaches \cite{58_fedavg}, FL typically trains a local model on each client and then aggregating the updated parameters in a central server. Directly applying this approach to train large-scale T2I models would require extensive computing resources and communication bandwidth on distributed client devices, which is infeasible. 
To address the resource and efficiency challenges in federated fine-tuning of large generative models, federated parameter-efficient fine-tuning methods \cite{7, 8, 9, 50_FedDAT} have been proposed, which freeze large portions of pre-trained models and update only selected layers. However, clients still need to locally host large-scale models. Split-training strategies \cite{10} offload computations to servers at the expense of communication costs and privacy risks. Another line of research \cite{11, 34_FedMKT} establishes cross-silo knowledge transfer between client small  models (SM) and a server large language models (LM). However, existing approaches are limited to unimodal clients. 

With the evolution of mobile intelligence and the Internet of Things (IoT) \cite{37}, a large volume of multi-modal data has been generated. Integrating these multi-modal data shows great promise \cite{38, 39, 40}. Yet effectively utilizing multimodal data in FL remains an open challenge.
In such environments, the heterogeneity in data modality, distributions, and model architectures makes direct knowledge alignment and fusion extremely difficult. 
Existing multimodal FL approaches rely on public datasets to unify multimodal data from various origins into consistent and aligned representations \cite{21_creamFL, 22_fedmekt}. 
However, the availability of public datasets in specialized fields is often constrained \cite{33}. In addition, previous work mainly focuses on modality alignment, overlooking the seamless integration of multimodal knowledge, a crucial aspect for T2I model fine-tuning. 
Our solution to these challenges is FedMMKT, a novel framework enabling privacy-preserving collaborative training and mutual enhancement between heterogeneous multimodal clients and a server T2I model via efficient cross-silo knowledge transfer. FedMMKT synthesizes domain-specific cross-modal data at the server (eliminating public data needs), leverages client ensemble knowledge to improve synthetic data quality and fuse multimodal representations, and uses these to fine-tune the T2I model and enhance clients. Extensive evaluations demonstrate significant performance improvements for  both clients and the server.

\section{Related Works}
\label{Related_Works}
\subsection{Multimodal Federated Learning}
Recent work in multimodal FL explores the integration of diverse modalities from decentralized clients to train a global multimodal model~\cite{25,26,27}. Key challenge is to effectively fuse knowledge from heterogeneous multimodal clients. Methods like MAML~\cite{18} and FedSea~\cite{19_fedsea} perform alignment over feature spaces, whereas FedMEKT~\cite{22_fedmekt} utilizes knowledge distillation. FedCola~\cite{Fedcola} addresses this challenge by collaboratively training a unified multimodal Transformer model through parameter aggregation. However, these approaches rely on the same client model architecture for each modality, neglecting the nuances of heterogeneous model structures. To address this, CreamFL~\cite{21_creamFL} introduces contrastive representation learning, and FedMBridge~\cite{62_fedmbridge} applies a topology-aware hypernetwork for architectural and statistical heterogeneity. DisentAFL~\cite{63_disentanglement} further disentangles inter-client knowledge flows via a two-stage transfer scheme. FedMMKT, differing from these predecessors, utilizes synthesized domain-specific data generated by large-scale models, securely incorporating client-side expertise to effectively fuse multimodal knowledge into a unified representation. This enables FedMMKT to enhance both server and client models while preserving data privacy and optimizing adaptability in federated learning scenarios.

\subsection{Fine-tuning T2I Models with Domain-Specific Data}
Pre-trained T2I models often underperform on domain-specific tasks and require further fine-tuning using specialized data~\cite{23,35,51_DomainStudio,52_dreambooth}. To reduce training cost, lightweight adapters~\cite{53_t2i} and training-free techniques~\cite{54_diffusiongpt} have been proposed. Human preference alignment is further improved via feedback-based methods~\cite{42,43_dreamsync}. Distinct from these approaches, FedMMKT fine-tunes T2I models without direct access to private data, instead leveraging synthetic domain-specific samples enriched by client-side knowledge.

\subsection{Federated Fine-tuning of Pre-trained Models}
Federated adaptation of large models addresses privacy and efficiency challenges in decentralized settings~\cite{zhuang2024foundationmodelmeetsfederated}.\textit{Parameter-based Approaches} uses parameter-efficient fine-tuning (PEFT) techniques \cite{zhang-etal-2023-fedpetuning}, such as soft prompts \cite{56_fedsp,59_promptfl,60_pfedprompt}, and adapters \cite{55_offsite,50_FedDAT,wu2024fedbiotllmlocalfinetuning,61_FlexLoRA} to address the communication and computation challenges. \textit{Distillation-based Approaches} \cite{34_FedMKT,31_fedgems,21_creamFL, peng2024fedpftfederatedproxyfinetuning} leverage knowledge distillation techniques \cite{57_distilling} to distill knowledge between a server LM and client small models. Finally, \textit{Generation-based Approaches} leverages the generative capabilities \cite{11,Zhang_2024_CVPR} of LMs to generate task-specific datasets to transfer knowledge.  However, most existing approaches remain restricted to single-modality tasks, limiting their applicability in integrating multimodal knowledge from heterogeneous clients. 

\section{FedMMKT Framework} 
\subsection{Problem Statement}
We consider a heterogeneous FL setting involving a cloud central server and $K$ clients. Among them, $N$ clients handle image classification tasks $\{C_i\}_{i=1}^N$, and the remaining $K-N$ clients handle text classification tasks $\{C_t\}_{t=1}^{K-N}$. Each client $k$ trains a private model $\mathcal{M}_k$  on its local dataset $D_k=\{(x_i,y_i)\}_{i=1}^{|D_k|}$  by minimizing:

\begin{equation}
L_{k} = \sum_{i=1}^{|D_k|} l(\mathcal{M}_k(x_i),y_i),\label{eq1}
\end{equation}
where $l$ is a standard loss (e.g., cross-entropy). A pre-trained T2I model $\mathcal{M}_{T2I}$ (with parameters $\Phi_{\mathcal{M}}$) resides on the server and is fine-tuned. To safeguard data privacy and model ownership , client data and local models do not leave their sites, and the server T2I model does not leave the server. The goal is to enhance both the T2I model’s generative ability and client model performance in domain tasks through privacy-preserving cross-modal knowledge transfer.

\subsection{Overview}
\begin{figure*}[ht]
  \centering
  \includegraphics[width=1.0\linewidth]{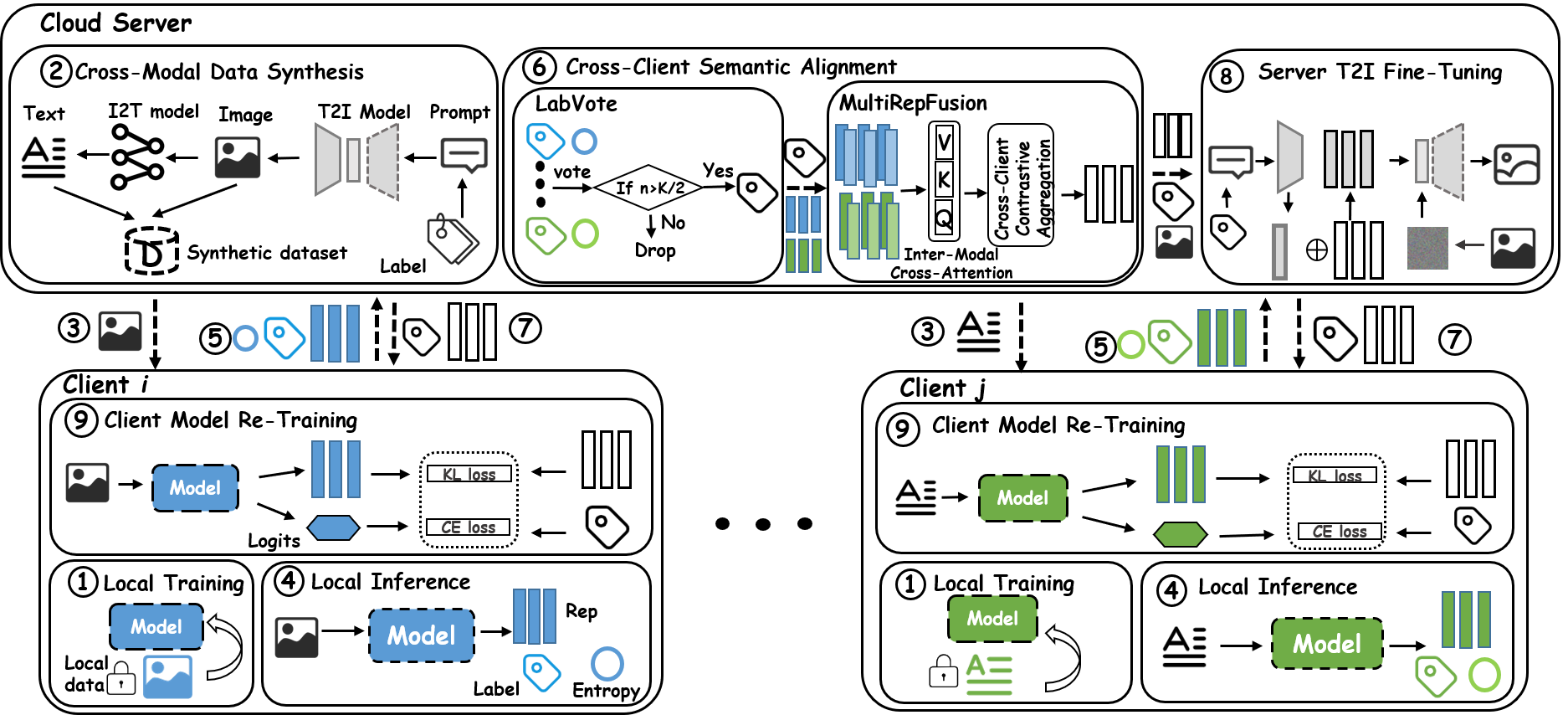}
  \caption{The workflow of the FedMMKT. Different colors represent different data modalities, while varying shades of each color indicate data originating from different clients; dash-dotted and solid borders denote updating and frozen components, respectively.The T2I model contains an encoder, a synthetic latent space, and a decoder.}
  \label{Fig.2}
\end{figure*}

The FedMMKT framework is illustrated in Fig.\ref{Fig.2}. The full algorithm in algorithm.\ref{alg:FedMMKT}. In a nutshell, FedMMKT first performs Cross-Modal Data Synthesis on the server to generate cross-modal data related to specialized domains, which are used for knowledge transfer between clients and the server (\ref{CrossSyn}); Then it intelligently fuses specialized knowledge from different clients by performing a novel ensemble label correction and multimodal representation fusion(\ref{CCSA}); Finally, the server T2I model and client models are fine-tuned using synthesized data and extracted multimodal representations (\ref{CS-BKT}). While designed for multimodal FL, FedMMKT can be adapted to unimodal and logit-based variants (see Appendix \ref{Appendix D} for details).

\begin{algorithm}
\caption{The Framework of FedMMKT}
\label{alg:FedMMKT}
\begin{algorithmic}[1]
\REQUIRE Number of communication rounds $T$, number of clients $K$, server model $\mathcal{M}_{T2I}$, dataset $D_k$ of the $k$-th client, label set $Y$, server fine-tuning epochs $E_s$, client re-training epochs $E_c$
\ENSURE $\mathcal{M}_k$, $\mathcal{M}_{T2I}$

\STATE \textbf{ClientLocalTraining:}
\STATE // Step 1: Initial Client Pre-training (once)
\FOR{client $k=1, ..., K$}
    \STATE $M_k \leftarrow ClientLocalTraining(D_k)$  \hfill $\triangleright$ Eq.~\ref{eq1}
\ENDFOR

\FOR{each round $t=1, 2, ..., T$}
    \STATE \textbf{ServerExecute:}
          \STATE // Step 2: Cross-Modal Data Synthesis
          \STATE $D^s \leftarrow SyntheticDatasetGeneration(Y)$\hfill $\triangleright$ Eq.~\ref{eq2}
          \STATE Distribute $D^s$ to all clients

    \STATE \textbf{ClientInference:}
    \STATE // Step 4: Local Inference on Clients
    \FOR{client $k=1, ..., K$}
         \STATE Local inference according to Eq.~\ref{eq3}
         \STATE Send $Y^{(k)}$, $E^{(k)}$ and representations $RI^{(k)}$ or $RT^{(k)}$ to server
    \ENDFOR

    \STATE \textbf{ServerExecute:}
    \STATE // Step 6: Cross-Client Semantic Alignment
    \STATE $D^s \leftarrow LabVote(I^s, T^s, Y^{(k)}, E^{(k)})$ \hfill $\triangleright$ Eq.~\ref{eq4}-\ref{eq5}
    \STATE Inter-Modal Cross-Attention according to Eq.~\ref{eq6}-\ref{eq7}
    \STATE Cross-Client Contrastive Aggregation according to Eq.~\ref{eq8}-\ref{eq10}
    \STATE Send $(Y^{s}, MR^{s})$ back to the clients

    \STATE // Step 8: Server T2I Fine-Tuning (for $E_s$ epochs)
    \STATE $M_{T2I} \leftarrow T2IFineTuning(Y^{s}, I, MR^{s})$ \hfill $\triangleright$ Eq.~\ref{eq11}-\ref{eq12}

    \STATE \textbf{ClientLocalTraining:}c
    \STATE // Step 9: Client Model Re-Training  (for $E_c$ epochs)
    \FOR{client $k=1, ..., K$} 
         \STATE $M_k \leftarrow ClientTraining(D^s, MR^{s})$ \hfill $\triangleright$ Eq.~\ref{eq13}
    \ENDFOR

  \STATE Return $M_k$, $M_{T2I}$
\ENDFOR
\end{algorithmic}
\end{algorithm}

\subsection{Cross-Modal Data Synthesis}
\label{CrossSyn}

Because private data are not directly accessible, FedMMKT first leverages the generative capabilities of the T2I model $\mathcal{M}_{T2I}$ to synthesize a domain-specific dataset $D^s$, which are used to transfer knowledge between clients and the server. 
Specifically, FedMMKT first randomly selects a class label $y_i$, converts it into a label-descriptive prompt $P(y_i)$, e.g., "a picture of $\{y_i\}$" (See Appendix \ref{Appendix B}), and then sent the prompt to $\mathcal{M}_{T2I}$ to synthesize the corresponding image $I_i^s$. 
\begin{equation}
  I_{i}^s \sim \mathcal{M}_{T2I}(\cdot|P(y_{i}),\Phi_{\mathcal{M}}) \, 
 \label{eq2}
\end{equation}

In order to fully utilize data from both images and text domains, we further obtain the textual descriptions corresponding to each generated image using a pre-trained image-to-text generation model (e.g., Blip \cite{24_blip} )'s API, denoted as $T_i^s$. Note that $T_i^s$ produced by an I2T model offers a much richer description of the image content, capturing specific characteristics and context (See Appendix \ref{Appendix B}). Afterwards, the constructed synthetic dataset $D^s=\{(I_i^s,T_i^s)\}_{i=1}^{|D^s|}$ is distributed to the clients according to modality, i.e. $T^s$ to text clients and $I^s$ to image clients. However, due to limited domain priors, synthetic images often fail to align with their input labels~\cite{2,41,44}. Especially, the images generated may not correspond correctly to its input labels (See Appendix \ref{Appendix C} for examples). We address this challenge by incorporating client-side supervision to correct and enhance the synthetic data, as described in Section~\ref{CCSA}.

\subsection{Cross-Client Semantic Alignment}
\label{CCSA}

\begin{figure*}[ht]
  \centering
  \includegraphics[width=1.0\linewidth]{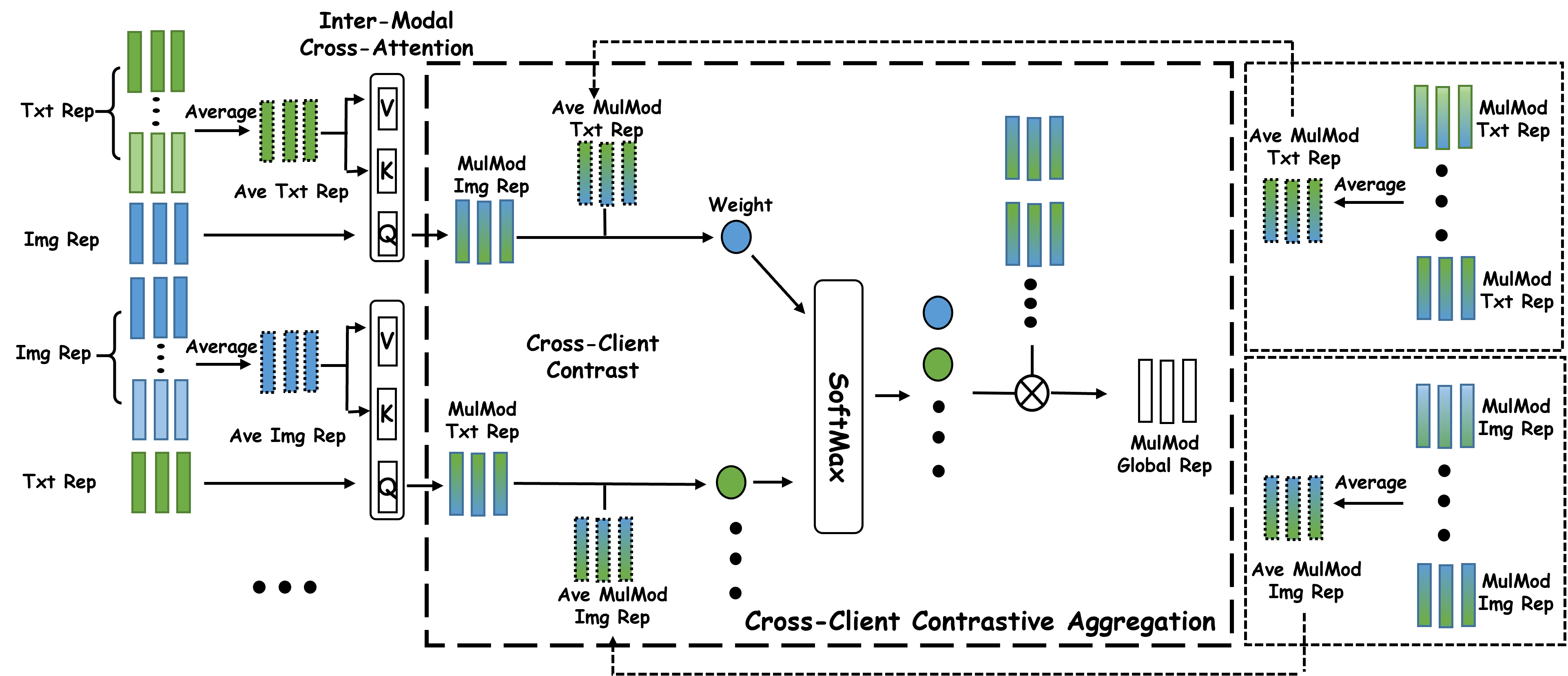}
  \caption{Workflow of MultiRepFusion.} 
  \label{Fig.3}
\end{figure*}
Upon receiving the synthetic dataset from the server, each client performs local inference using its model $\mathcal{M}_k$, which is decomposed into a feature extractor $f_{\theta_k}$ parameterized by $\theta_k$ and a classifier $g_{\phi_k}$. For each synthetic image $I_i^s$, the client extracts a representation \(RI_i^{(k)} = f_{\theta_k}(I_i^s)\)and predicts class probabilities $p_i^{(k)}$ over the label space $\mathcal{T}$. The predicted label is given by \(Y_i^{(k)} = \arg\max_{t \in \mathcal{T}} p_i^{(k)}(t)\), and the prediction confidence is measured by the entropy:

\begin{equation}
E_i^{(k)} = \frac{1}{1 - \sum_{t \in \mathcal{T}} p_i^{(k)}(t) \log p_i^{(k)}(t)}.
\label{eq3}
\end{equation}

A lower entropy corresponds to a higher weight $E_i^{(k)}$, indicating greater confidence in the prediction. Text clients process synthetic texts analogously to obtain $RT_i^{(t)}$, $Y_i^{(t)}$, and $E_i^{(t)}$.

Since each client trains its local model using its own private data, the representations generated by different clients tend to diverge significantly \cite{11}, posing substantial challenges for effectively aligning and integrating them into a coherent global representation. In order to effectively integrate specialized multimodal domain knowledge from diverse clients, FedMMKT introduces two mechanisms: LabVote, which uses entropy-weighted voting to correct noisy labels and filter unreliable samples; and MultiRepFusion, which integrates diverse client representations via inter-modal attention and contrastive learning to form a rich, unified embedding for downstream transfer. 

\subsubsection{LabVote} To address the low-quality issue in \ref{CrossSyn}, we leverage client knowledge to correct labels of the synthetic data. This correction is pivotal for enhancing the effectiveness of subsequent training. Each client’s prediction $Y_i^{(k)}$ is weighted by its confidence $E_i^{(k)}$, and the consensus label is computed as:
\begin{equation}
Y_i^{s} = \arg\max_{y \in \mathcal{Y}} \left\{\sum_{k=1}^{K} E_i^{(k)} \cdot \delta(Y_i^{(k)} = y)\right\}. 
\label{eq4}
\end{equation}

To ensure the quality and reliability of the synthetic data, a filtering mechanism is further applied. Specifically, if the label $Y_i^{s}$ has fewer votes than $\beta$ of total clients ($0 \leq \beta \leq 1$), the $i$-th data is considered to be unreliable and is discarded.
\begin{equation}
D^{s'} = \{(I_i^s, T_i^s, Y_i^s)\}_{v_i > \lfloor\beta{K}\rfloor},   \ \ 
\label{eq5}
\end{equation}
where $v_i$ denotes the final votes for $Y_i^s$. In this work, we chose $\beta =0.5$ based on empirical evaluations. We found this threshold is effective in maintaining a high level of label accuracy while preserving sufficient data diversity for the subsequent knowledge fusion process. The refined high-quality data $D^{s'}$ are then fed into next module for multimodal feature fusion.

\subsubsection{MultiRepFusion} The main goal of MultiRepFusion is to fuse representations from different clients and modalities to facilitate knowledge transfer between clients and the server. Fig.\ref{Fig.3} shows the workflow. Due to the distinct nature of the data and tasks processed by each client, the uploaded representations inevitably exhibit two primary types of bias: \textit{modality bias} and \textit{distribution bias}. \textit{Modality bias} arises because image clients have no exposure to textual information during training, while text clients lack visual data. This separation of modalities leads to substantial discrepancies in the representations generated for the same synthetic data. \textit{Distribution bias} stems from the non-independent and identically distributed (non-IID) nature of data across clients, causing the representations learned for the same synthetic data to diverge and intensifying inconsistencies among clients. To address the representation discrepancies from different clients, the following two steps are taken. 

\textbf{Inter-Modal Cross-Attention.} 
To address the \textit{modality bias} suffered by uni-modal clients, FedMMKT first employs inter-modal cross-attention for multimodal representation alignment. Specifically, inter-modal cross-attention is applied to integrate local unimodal representations with complementary information from the other modality, represented by the averaged local representations from clients of the other modality:

\begin{equation}
MRI_i^{(n)} = \text{CA}\left(RI_i^{(n)}, \frac{\sum_{t=1}^{K-N} RT_i^{(t)}}{K-N}\right)
\label{eq6}
\end{equation}
\begin{equation}
\label{eq7}
MRT_i^{(t)} = \text{CA}\left(RT_i^{(t)}, \frac{\sum_{n=1}^{N} RI_i^{(n)}}{N}\right) 
\end{equation}

where CA represents cross-attention. Averaging over all clients can mitigate the impact of client-specific discrepancies, emphasizing common patterns across all clients, thereby providing a more balanced and consistent view of the other modality.  With this cross-attention mechanism, different modalities are effectively integrated, and local representations are enriched by the complementary knowledge from the other modality. This step reduces modality-specific biases and enables more coherent multimodal knowledge transfer.

\textbf{Cross-Client Contrastive Aggregation.} After representation alignment, we aim to fuse the multimodal representations from all clients into a unified representation for subsequent knowledge transfer. To address the \textit{distribution bias} among clients, we design a cross-client, cross-modal contrastive score for weighting clients. Specifically, a representation that is "closer" to the average representation of its paired counterpart while being "further" away from other unrelated samples, is considered to better capture the underlying semantic information. Consequently, the representation weight $w_i^{(n)}$ of the $i$-th data of the $n$-th client is calculated as follows:

\small
\begin{equation}
w_i^{(n)} = \log \left(\frac{\exp\left(\scriptstyle{\text{Cos}(MRI_i^{(n)}, \frac{\sum_{t=1}^{K-N} MRT_i^{(t)}}{K-N})}\right)}{\sum_{j=1}^{\text{batch}} 1_{[j \ne i]} \exp\left(\scriptstyle{\text{Cos}(MRI_i^{(n)},\frac{\sum_{t=1}^{K-N} MRT_j^{(t)}}{K-N})}\right)}\right) \label{eq8}
\end{equation}
\normalsize
where, $\text{Cos}(u,v) = \frac{u^\top v}{\|u\|\|v\|}$ denotes the cosine similarity. Next, the server normalizes and aggregates the client-side representations as follows:
\begin{equation}
\alpha_i^{(1)}, \alpha_i^{(2)}, \ldots, \alpha_i^{(K)} = \text{softmax}(w_i^{(1)}, w_i^{(2)}, \ldots, w_i^{(K)}) \label{eq9}
\end{equation}
\begin{equation}
MR_i^{s} = \sum_{n=1}^N \alpha_i^{(n)} \cdot MRI_i^{(n)} + \sum_{t=1}^{K-N} \alpha_i^{(t)} \cdot MRT_i^{(t)}\label{eq10}
\end{equation}

Note unlike CreamFL\cite{21_creamFL}, we aggregate both the image representations and text representations into a unified representation. In addition, we contrast with the averaged representations from the other modality, while CreamFL contrasts with a global representation from the server model. These approaches are taken for enabling subsequent T2I model fine-tuning. 

\subsection{Client-Server Knowledge Transfer}
\label{CS-BKT}

Finally, fused representations are merged into $D^{s'}$, i.e., $D^{s'}=\{(MR_i^{s}, {Y}_i^{s}, I_i^s,T_i^s)\}_{i=1}^{|D^{s'}|}$, which are utilized by the server and clients to update their respective models.

\subsubsection{Server T2I Fine-Tuning}
The original T2I model consists of an encoder $e(\cdot)$ and a decoder $d(\cdot)$. Inspired by PITI \cite{36}, we integrate the aggregated global representation with textual embeddings from the encoder, formulated as: 
\begin{equation}
    z = MR_i^{s} + e(P(Y_i^{s}))\label{eq11}
\end{equation}
The decoder $d(\cdot)$ then takes $z$ as input to generate images that embody the intended meaning of the text prompt and the integrated mutimodal representation. The primary training objective here is to optimize the decoder component $d(\cdot)$  while keeping the encoder and other components fixed. This ensures stable latent representation extraction while focusing on improving image generation. It also ensures that the reconstructed image remains closely aligned with the conditions set by the provided multimodal inputs, thus enhancing the contextual accuracy of the generated output. 

In order for the T2I model to perform independent inference without multimodal representations from clients, we introduce a stochastic component during fine-tuning, inspired by \cite{1} . Specifically, we employ a Bernoulli random variable $B(p)$, where $p$ represents the probability of omitting image representations as follows:
\begin{equation}
    z = 
    \begin{cases} 
          MR_i^{s} + e(P(Y_i^{s})) & \text{with probability } 1 - p \\
          e(P(Y_i^{s})) & \text{with probability } p 
    \end{cases}
    \label{eq12}
\end{equation}
In this work, we set $p=0.2$. This stochastic approach ensures that the model maintains its generative capabilities even when multimodal inputs are unavailable.

\subsubsection{Client Model Re-Training}
On the client side, the local models trained on private data are also subject to bias due to the non-IID distribution and missing modality.  To address these biases, we incorporate refined synthetic data with fused representations $D^{s'}$to re-train their local models  to obtain an enhanced model $\mathcal{M'}_k$. 
Specifically, we encourage the local model to align its representations with the fused representations with a Kullback-Leibler (KL) divergence loss, and align its local predictions with the consensus label.

\begin{equation}
L_{\text{local}} = \sum_{i=1}^{|D^{s'}|} \left[ D_{\text{KL}}\left(R_i^{(k)} \parallel MR_i^s\right) + \lambda \cdot l\left(\mathcal{M'}_k(x_i^s), Y_i^s\right) \right]
\label{eq13}
\end{equation}
\section{Theoretical Analysis}
\subsection{Privacy-Preserving Discussion}
\label{Privacy-Preserving Discussion}
In FedMMKT, privacy is preserved from both data and model aspects. Firstly, only synthesized multimodal data is transmitted during training, ensuring data privacy and minimizing risks of leakage. Secondly, all model parameters remain local and are never shared, safeguarding client model ownership and prevent data leakage from model inversion attacks. Following its predecessors\cite{22_fedmekt, 21_creamFL}, our framework does not treat labels ${y_i}$ as sensitive. Additional privacy-preserving techniques such as Differential Privacy \cite{65_differential} can be applied to enhance label confidentiality.

\subsection{Convergence Analysis}

In this section, we provide convergence guarantees for FedMMKT under commonly adopted assumptions. The algorithm proceeds by having each client $k$ minimize its local objective in each round, as described in Eq.~\ref{eq12}. Before formally stating the convergence results, we introduce some notations.

Let $x$ denote the input data and $f_{\theta}$ the feature extraction function. $L$ denotes the global objective, which is the weighted sum of local objectives $L_k$. For the sake of the convergence proof, we assume the following key conditions:

\textbf{Assumption 1 (Smoothness)}: The local objective $L_k(w)$ is $L$-smooth for all clients:
\begin{equation}
\|\nabla L_k(w) - \nabla L_k(w')\| \le L \|w - w'\|, \quad \forall w, w' \in \mathbb{R}^d
\label{eq20}
\end{equation}

\textbf{Assumption 2 (Bounded Variance)}: We assume that the variance of the local gradients is bounded. Specifically, for all clients, the following holds:
\begin{equation}
\frac{1}{K}\sum_{k=1}^{K} \|\nabla L_k(w) - \nabla L(w)\|^2 \le \zeta^2
\label{eq21}
\end{equation}
where $\zeta$ quantifies the non-IID degree of the data across clients. 

\textbf{Assumption 3 (Client Drift)}: We assume that the drift of the client models during local updates is bounded:
\begin{equation}
\mathbb{E}\|w_t^{(k)} - w_t\|^2 \le \gamma^2
\label{eq22}
\end{equation}
where $\gamma^2$ represents the deviation of the local model updates from the global model, reflecting the effects of local optimization steps and learning rates. 

\textbf{Assumption 4 (Multi-modal Alignment Error)}: In FedMMKT, due to the heterogeneous nature of the data across different modalities, there is an alignment error between the local representations and the global fused representation. Specifically, we assume:
\begin{equation}
\mathbb{E}\|R^{(k)} - MR^s\|^2 \le \epsilon_{\text{align}}^2
\label{eq23}
\end{equation}
where $R^{(k)}$ represents the representation from client $k$, and $MR^s$ is the aggregated multi-modal representation on the server.
Following prior works such as FedProx \cite{FedProx}, FedGKD \cite{FedGKD}, and FedSSD \cite{FedSSD}, we introduce a lemma to justify the rationality of the surrogate objective.

\textbf{Lemma 1:} Let $\tilde{L}(w; w_t)$ be the local optimization objective:
\begin{equation}
\tilde{L}(w; w_t) = \sum_{i=1}^{|D|} \left[ \sigma_{\text{KL}} L_{\text{KL}}^2(w, w_t) + \lambda \sigma_{\ell} L_{\ell}^2(w, w_t) \right],
\label{eq24}
\end{equation}
 
where $L_{\text{KL}}^2(w, w_t)$ represents the KL divergence term; $L_{\ell}^2(w, w_t)$ represents the cross-entropy term; $\sigma_{\text{KL}}$ and $\sigma_{\ell}$ are constants related to the smoothness of the losses; $\lambda$ is the regularization factor.
 
We know that both KL divergence and cross-entropy loss are Lipschitz continuous under mild conditions, as shown in prior works~\cite{FedProx,FedGKD,FedSSD}. Hence, by using standard Lipschitz continuity assumptions, we can bound the objective function. This gives us:
\begin{equation}
\tilde{L}(w; w_t) \leq L(w_t, w_t) = L(w_t)
\label{eq25}
\end{equation}
 This inequality shows that the surrogate objective $\tilde{L}$ is a non-increasing function over iterations. 
 
\textbf{Proof:}
Since both KL divergence and cross-entropy loss are Lipschitz continuous, the terms $\sigma_{\text{KL}} L_{\text{KL}}^2(w, w_t)$ and $\sigma_{\ell} L_{\ell}^2(w, w_t)$ satisfy the following inequalities:
\begin{equation}
\begin{split}
&\|\nabla L_{\text{KL}}(w) - \nabla L_{\text{KL}}(w')\|\\
& \leq L_{\text{KL}} \|w - w'\|
\|\nabla L_{\ell}(w) - \nabla L_{\ell}(w')\|\\
& \leq L_{\ell} \|w - w'\|    
\end{split}
\label{eq26}
\end{equation}
Thus, we can conclude that the surrogate objective $\tilde{L}(w; w_t)$ is non-increasing at each iteration. This ensures that the global objective function $L(w)$ is non-increasing as well, proving the validity of the surrogate bound. 

\textbf{Theorem 1 (Non-convex Convergence):} Let the assumptions above hold. Then, the following inequality holds for FedMMKT:
\begin{equation}
\begin{split}
\min_{t \in \{0,\ldots,T-1\}} \mathbb{E}\|\nabla L(w_t)\|^2 
\le &\frac{2(L(w_0)-L^*)}{\eta T (1-\beta)} \\
&+ 
O(\eta L^2 (\zeta^2 + \gamma^2 + \epsilon_{\text{align}}^2))
\end{split}
\label{eq23}
\end{equation}
where $L^*$ denotes the minimum global objective value, and $\beta \in [0, 1)$ is the effective curvature parameter. Specifically, $\beta$ arises from the proximal-like term in the local objective, analogous to FedProx~\cite{FedProx}, ensuring
\[
\tilde{L}(w_{t+1}; w_t) \le \tilde{L}(w_t; w_t),
\]
where $\tilde{L}$ is the regularized objective in each communication round. This term controls the stability of local updates under heterogeneous conditions.

This result shows that the global objective decreases over time, and the algorithm converges to a stationary point. The convergence rate is $O(1/T)$, which is typical for non-convex problems under appropriate conditions.

\textbf{Proof:}
From \textbf{Lemma 1}, we know that the surrogate objective $\tilde{L}(w_t)$ is non-increasing:
\begin{equation}
\tilde{L}(w_{t+1}; w_t) \leq \tilde{L}(w_t; w_t)
\label{eq24}
\end{equation}
This implies that the global objective $L(w)$ decreases over time:
\begin{equation}
L(w_{t+1}) \leq L(w_t)
\label{eq25}
\end{equation}
Since the objective is $L$-smooth (\textbf{Assumption 1}), we have the following gradient descent inequality:
\begin{equation}
\|\nabla L(w_{t+1})\|^2 \leq \|\nabla L(w_t)\|^2 - \frac{1}{\eta} (L(w_{t+1}) - L(w_t))
\label{eq26}
\end{equation}
Using the gradient descent update rule, we can further deduce:
\begin{equation}
L(w_{t+1}) \leq L(w_t) - \eta \|\nabla L(w_t)\|^2 + O(\eta^2)
\label{eq27}
\end{equation}
From \textbf{Assumption 2} and \textbf{Assumption 3}, we know that the variance of the gradients and the client drift are bounded by $\zeta^2$ and $\gamma^2$, respectively. Thus, the global gradient variance and client drift contribute an error term:
\begin{equation}
\frac{1}{K} \sum_{k=1}^{K} \|\nabla L_k(w) - \nabla L(w)\|^2 \leq \zeta^2, 
\label{eq28}
\end{equation}
\begin{equation}
\quad \mathbb{E} \|w_t^{(k)} - w_t\|^2 \leq \gamma^2.
\label{eq29}
\end{equation}
Combining these factors and \textbf{Assumption 4}, we obtain \textbf{Theorem 1}.

\textbf{Discussion.} The convergence analysis shows that, under the assumption of smoothness and bounded variance, FedMMKT achieves convergence to a stationary point with a convergence rate of $O(1/T)$. The error terms indicate how different factors impact the final convergence:

- The **non-IID term** $\zeta^2$ reflects the effect of heterogeneous data across clients. As expected, when clients have very different data distributions, convergence is slower. 

- The **client drift term** $\gamma^2$ captures the impact of local updates and client-specific optimization steps. Larger values of $\gamma$ result in slower convergence, but FedMMKT can mitigate this effect through knowledge transfer techniques like cross-client aggregation. 

- The **multi-modal alignment error** $\epsilon_{\text{align}}^2$ shows the challenges of aligning knowledge across different modalities (e.g., images and text). FedMMKT incorporates mechanisms like cross-attention and contrastive aggregation to reduce this error.

The convergence rate of FedMMKT is influenced by the degree of data non-IIDness and multi-modal misalignment. If these factors are large, the convergence rate may be slower due to the increased variance and the challenges in fusing multi-modal knowledge. However, the algorithm is still guaranteed to converge to a first-order stationary point, as shown in Theorem 1.

\textbf{Corollaries.}
\paragraph{Linear Speedup w.r.t. Communication Rounds} 
If $\zeta, \gamma, \epsilon_{\text{align}} = O(T^{-1/2})$, then FedMMKT achieves the standard $O(1/T)$ convergence rate for non-convex optimization.
\paragraph{Bounded Multimodal Bias}  If the cross-modal fusion module satisfies:
\[
\epsilon_{\text{align}}^2 \le \frac{c}{L^2}
\]
for some constant $c < 1$, then the global error bound of FedMMKT can be further reduced to:
\[
O\left(\frac{1}{T} + \zeta^2 + \gamma^2\right).
\]
\paragraph{Communication Complexity}
Beyond the convergence rate above, the number of communication rounds required to achieve $\min_{t}\mathbb{E} \,\|\nabla L(w_t)\|^2 \le \varepsilon$ is
\[
T = \tilde{O}\!\left( \frac{1}{\varepsilon^2}\bigl(\sigma^2 + L^2(\zeta^2 + \gamma^2 + \epsilon_{\text{align}}^2) \bigr) \right),
\]
where $\sigma^2$ is the variance bound of stochastic gradients. This quantifies the tradeoff between accuracy and the cost in federated communication.
\paragraph{Client Drift Control}
The drift term $\gamma^2$ is reduced in FedMMKT by LabVote and MultiRepFusion modules:  
(i) \emph{LabVote} enforces consistency in pseudo-label generation across clients, acting as a soft regularizer;  
(ii) \emph{MultiRepFusion} distills aligned multi-modal knowledge back to clients, reducing local parameter divergence from the global model.

In conclusion, the theoretical analysis of FedMMKT under the above assumptions provides convergence guarantees in non-convex settings, even with non-IID data distributions and multi-modal misalignments. By incorporating the effects of client drift and modality alignment errors, we gain a more realistic understanding of the algorithm's convergence behavior. The results suggest that FedMMKT can converge effectively while accounting for the challenges posed by heterogeneous data and the integration of multi-modal knowledge.

\section{Experiments}
\label{Experiments}
\subsection{Setup}
\label{Setup}
\textbf{Datasets and Models.} We evaluated FedMMKT on Oxford 102 Flower \cite{16_flowers102} and UPMC Food-101 \cite{17_food101} datasets. For each dataset, we partitioned the training data across different clients following a  Dirichlet distribution \((\alpha = 0.5)\) \cite{29_dl}. We utilized six different model architectures: ViT-B/32 \cite{12_vit}, BeiT \cite{13_beit}, and Resnet18 for image clients, BERT \cite{14_bert}, DistilBERT \cite{15_distilbert}, and RoBERTa \cite{64_roberta} for text clients. We consider two heterogeneous FL scenarios: 1) heterogeneous image clients only, and 2) heterogeneous image and text clients. The server employs GLIDE~\cite{1} or Stable Diffusion~\cite{3} as the T2I model, fine-tuned using Adam (initial learning rate 0.0002) with cosine annealing.

\textbf{Baselines.} 
We compared with heterogeneous FL methods including FedMD \cite{30_fedmd}, FedGems \cite{31_fedgems}, FedET \cite{32_fedet}, FedMEKT \cite{22_fedmekt}, CreamFL \cite{21_creamFL}, DisentAFL\cite{63_disentanglement}, and FedMBridge\cite{62_fedmbridge}. We additionally include several parameter-efficient fine-tuning approaches: LoRA-FL, Bias-FL, and FedDAT~\cite{50_FedDAT}. Among them, FedMEKT, CreamFL, DisentAFL, and FedMBridge are multi-modal FL algorithms. We also report two FedMMKT variants: u-FedMMKT, where clients use only a single modality, and l-FedMMKT, which uses logit-level instead of representation-level transfer (see Appendix~\ref{Appendix D}). 

\textbf{Metrics:} 
We use CLIP score \cite{46} to evaluate the alignment quality between the generated content and the textual prompts of T2I model. In addition, we adopt GAN-test\cite{45} to evaluate the performance on domain adaption, defined as T2I accuracy. 
While CLIP score offers a broad, generic evaluation of text-image alignment, the T2I accuracy provides a targeted assessment on domain-specific knowledge. 

\subsection{Performance}
\subsubsection{Performance of Client Models}
\begin{table*}[ht]
\centering
\caption{Performance comparison of client models with multimodal FL baselines. “Img Acc” and “Txt Acc” denote average accuracy for image and text clients; “Average” is the mean over all datasets and modalities. “Standalone” trains locally without knowledge transfer. “ViT+BeiT (2+2)” represents 2 clients with ViT and 2 with BeiT. GLIDE is used as the server T2I model for FedMMKT and its variants. CreamFL requires paired modalities and is not applicable in image-only settings.} 
\resizebox{0.85\linewidth}{!}
{
\begin{tabular}{ccccccccccc}
\toprule 
  \multirow{2}{*}{\textbf{Mode} }&  \multirow{2}{*}{\textbf{Client Model}}  &  \multirow{2}{*} {\textbf{Method}}  &\multicolumn{2}{c}{ \textbf{Flowers102}} &  \multirow{2}{*}{\textbf{Food101}} & \multirow{2}{*}{\textbf{Average}} \\
  &  & & \textbf{Img Acc} & \textbf{Txt Acc} & \textbf{Img Acc} & \textbf{Txt Acc}&  \\
\cline{1-8} 

 \multirow{6}{*}{Image clients only }& \multirow{6}{*}{ViT+BeiT(2+2)}   & Standalone & 65.4(1)  &-  & 66.2(2) & - &65.8\\
& & DisentAFL &  72.8(5)  & - & 71.2(8)& - &72.0 \\
                 &  & FedMBridge&73.7(8)  & - & 72.4(7) &  - &73.1\\
                &  & FedMEKT&74.0(9)  & - & 72.8(8) &  - &73.4\\
                &  & u-FedMMKT                      & \textbf{81.8(4)} & - & \textbf{79.8(5)}&       - & \textbf{80.1} \\
                &  & l-FedMMKT    & 79.7(3) & - &      78.8(5) &    -  & 79.3\\
\cline{1-8}

 \multirow{23}{*}{Image and text clients}&  \multirow{6}{*}{{\begin{tabular}[c]{@{}c@{}}ViT+BeiT:\\ BerT+DistilBerT\\(2+2:2+2)\end{tabular}}}  & Standalone & 65.4(1)  &66.8(3)  &  66.2(2) &65.9(3)  & 66.1\\
&                       & DisentAFL &75.3(6)  &68.1(7)  &73.6(8) & 67.1(9) &71.0\\  
     &              & FedMBridge   &76.1(4)   &68.8(5) &74.8(4) & 67.9(8) &71.9 \\
& & CreamFL                     &77.8(5)                 &69.6(4)  &  76.6(3) &      68.5(6)  &73.1    \\
& & FedMEKT                     &78.1(5)                &69.8(6)  &  77.3(5) &      68.9(4)    &73.5 \\
  & & FedMMKT                  & \textbf{84.1(5)} &                      \textbf{70.9(5)} &  \textbf{81.4(6) }    &   \textbf{70.0(5)}  &\textbf{76.6}  \\
 &  & l-FedMMKT                  & 81.6(4)               & 70.1(4)  & 80.6(4)   &     69.7(5)     &75.5    \\
\cline{2-8}
& \multirow{6}{*}{{\begin{tabular}[c]{@{}c@{}}ViT+BeiT+ResNet18:\\ BerT+DistilBerT+RoBERTa\\(2+2+2:2+2+2)\end{tabular}}}  & Standalone & 60.2(5)  &64.0(2) &  63.2(5) &62.5(5) &62.5 \\
&                        & DisentAFL &71.1(6)  &64.9(5)  &69.0(3) & 63.8(7) &67.2\\  
     &              & FedMBridge   &72.6(4)   &65.2(2) &70.2(3) &64.2(8) &68.1 \\
& & CreamFL                     &73.6(5)                 &65.8(4)  &  72.8(3) &      64.9(6)  &69.3    \\
& & FedMEKT                    &74.2(6)                &66.2(5)  &  73.5(6) &      65.4(5)   &69.8   \\
  & & FedMMKT                  & \textbf{79.0(5)} &                      \textbf{68.0(5)} &  \textbf{77.7(8)}    &   \textbf{67.1(5)}  & \textbf{72.9}  \\
 &  & l-FedMMKT                  & 77.0(4)                & 66.6(3)  & 76.6(5)   &     66.3(3)      &71.6   \\
\cline{2-8}
& \multirow{6}{*}{{\begin{tabular}[c]{@{}c@{}}ViT+BeiT:\\ BerT+DistilBerT\\ (4+4:4+4)\end{tabular}}}& Standalone  & 58.3(2)  &  60.2(2) &61.7(2)  &59.3(4) &59.9\\
& & DisentAFL &68.5(6) &61.8(8) &65.1(8) & 60.9(9) &64.1\\
                     &     & FedMBridge &70.4(3) &60.9(7)&  66.0(9) &61.2(7)&64.6\\
& & CreamFL       &71.9(8)  &61.9(3) &  68.1(7)    & 61.7(7 &65.9                                \\
& &FedMEKT       &72.1(6)  &62.2(6) &  68.6(7)    & 61.9(6)   &66.2                              \\
  & & FedMMKT      & \textbf{77.5(7)}         & \textbf{63.8(6)} & \textbf{69.9(6)}  &           \textbf{62.8(4)}       & \textbf{68.5}       \\   
 &  & l-FedMMKT    & 76.5(6)  &63.0(6)& 69.5(4)&   62.5(5)    &67.9   \\
\bottomrule
\end{tabular}
}
\label{table1}
\end{table*} 
In Table~\ref{table1}, we compare the performance of FedMMKT on the client with other multimodal FL baselines. Notably, only FedMMKT and its variants employ a server-side T2I model, while other methods are follow their original architectures, relying solely on client-side aggregation  or alignment mechanisms without any generative component. As shown in Table~\ref{table1}, all FL methods consistently outperform the "Standalone" baseline, demonstrating the benefits of cross-client knowledge sharing. Among all methods, FedMMKT achieves the best performance, improving accuracy by up to 18.7\% (image) and 4.6\% (text) on Flowers102, and 15.4\% and 4.9\% on Food101. These gains stem from its ability to synthesize high-quality data and integrate multimodal knowledge effectively. The performance gain of logit-based variant (l-FedMMKT) is less, validating the advantage of representation-level transfer. Moreover, the image clients' accuracy in the multimodal (2+2:2+2) setting outperforms the unimodal (2+2) case for all baselines, demonstrating the value of modality complementarity. Finally, although performance declines as heterogeneity increases for all methods, FedMMKT remains the most robust, with the smallest average performance drop (7.85\%) compared to other methods (8.05–8.50\%), showing strong scalability and adaptability in highly diverse FL environments. In the 'Image only' setting, the multimodal baselines were adapted by removing their text client branches. 

\subsubsection{Performance of T2I model}
\begin{table*}[ht]
\centering
\caption{Performance comparison of T2I with different fine-tuning strategies using GLIDE under ViT+BeiT:
BerT+DistilBerT (2+2:2+2) client setting.}
\resizebox{0.75\linewidth}{!}{
\begin{tabular}{cccccc}
\toprule
 \multirow{2}{*}{\textbf{Method} }& \multirow{2}{*}{\textbf{Data}}  &\multicolumn{2}{c}{\textbf{Flowers102} }&\multicolumn{2}{c}{\textbf{Food101}}\\
  &  & \textbf{T2I Acc} & \textbf{Clip score}& \textbf{T2I Acc} & \textbf{Clip score}  \\
 \cline{1-6}
-&None 
 & 67.1(5) & 0.2975(5) & 60.4(9)  & 0.2847(5) \\
\cline{1-6}
\multirow{3}{*}{Self-fine-tuning}&\(I^s + Y^s\) & 72.4(5)  & 0.3010(4)& 64.9(8)  & 0.2891(8)\\
&\(I^s + T^s + Y^s\) &73.6(7) &0.3020(5) &67.8(6)  & 0.2920(5)\\
&\(I^s + Y^s + Rep\)& 72.6(8)&0.3012(5) & 66.5(7)  & 0.2907(8)\\
\cline{1-6}
l-FedMMKT
&\(I^s +T^s + Y^{s'}\) & 77.0(6) &0.3055(3)   & 75.1(5) & 0.2985(3) \\
u-FedMMKT&\(I^s + Y^{s'} + MRI\)  & 77.5(5) &0.3057(5)  & 72.8(4) & 0.2966(4)\\
FedMMKT&\(I^s + T^s + Y^{s'} + MRI + MRT\)  & \textbf{81.2(7)} &\textbf{ 0.3083(2)}  & \textbf{77.2(6)} & \textbf{0.3004(4)}\\
\cline{1-6}
Centralized fine-tuning & Private data
 & 87.2(3) & 0.3133(2) & 85.3(9)  & 0.3096(4)\\
\bottomrule
\end{tabular}
}
\label{table2}
\end{table*}

Table~\ref{table2} compares T2I model performance under four fine-tuning strategies: no fine-tuning,  self-fine-tuning using its generated data only without knowledge transfer from clients,FedMMKT and l-FedMMKT, and centralized fine-tuning with private data (“Private data”), with the last one serving as the performance upper bound. Other FL baselines are not included, as they do not consider server-side T2I fine-tuning. Within “self-fine-tuning”, combining synthetic images with generated text ("\(I^s + T^s + Y^s\)") outperforms using image-label pairs ("\(I^s + Y^s\)"), highlighting the benefit of multimodal information. Additionally, "\(I^s + Y^s + Rep\)," where representations were extracted from a pre-trained ViT model and used in T2I fine-tuning (using Eq.~\ref{eq11}),  offers limited improvement, suggesting that representations alone lack sufficient contextual grounding. FedMMKT yields substantial gains (~10\% on T2I accuracy) over self-fine-tuning baselines, demonstrating the effectiveness of knowledge transfer from private clients. 

\subsubsection{Performance comparison with the integration of a server T2I model}
\begin{table*}[ht]
\centering
\caption{Comparison of FedMMKT with baselines using GLIDE on Flowers102 dataset.}
\resizebox{0.65\linewidth}{!}
{
\begin{tabular}{ccccccc}
\toprule
 \textbf{Client Model} & \textbf{Transfer} & \textbf{Method} &  \textbf{T2I Acc} &\textbf{Clip score} & \textbf{Img Acc} & \textbf{Txt Acc} \\
\cline{1-7}

    \multirow{7}{*}{ViT+BeiT(2+2)} &- & Standalone   & - & -&  65.4(1)  & - \\
   
\cdashline{2-7}
                     &\multirow{4}{*}{Logit}  & FedET   & 75.7(5) & 0.3036(2)                        & 79.4(3) & - \\
                     &  & FedMD    & 74.5(6) & 0.3026(2)                        & 78.7(6) & - \\
                    &  & FedGems  & 75.5(4) & 0.3035(3)& 78.3(3) & - \\
                     &  & l-FedMMKT  & 76.2(4) & 0.3042(3)                      & 79.7(3) & - \\
                     \cdashline{2-7}
    & \multirow{2}{*}{Rep}  
                         & FedMEKT & 75.6(6) & 0.3037(4) & 79.6(6) & - \\
                     &  & u-FedMMKT  & \textbf{77.5(5)} & \textbf{0.3057(2)}& \textbf{81.8(4)} & -  \\
\cline{1-7}
   \multirow{7}{*}{{\begin{tabular}[c]{@{}c@{}}ViT+BeiT:\\ BerT+DistilBerT\\(2+2:2+2)\end{tabular}}} & - & Standalone   & - & -&  65.4(1)  &66.8(3) \\
   
\cdashline{2-7}        
                     &\multirow{4}{*}{Logit} 
                         
   & FedET    & 75.5(7) & 0.3038(6)&  79.0(6) &  69.5(7) \\
                    &  & FedMD    & 75.4(6) & 0.3037(3)& 79.2(5) &  69.2(5) \\
                     &  & FedGems  & 76.4(5) & 0.3046(2)&80.2(4)& 69.6(4) \\
                     &  & l-FedMMKT   & 77.0(6) & 0.3055(3)&81.6(4) & 70.1(4) \\
                     \cdashline{2-7}
 &  \multirow{3}{*}{Rep}   
                         & FedMEKT  & 76.5(5) & 0.3046(2)& 81.0(5)  & 69.4(4) \\
                     &  & CreamFL  & 77.0(9) & 0.3051(7)& 81.6(6) & 70.5(7) \\
                     &  & FedMMKT   & \textbf{81.2(7)} &\textbf{ 0.3083(2)}& \textbf{84.1(5)} &  \textbf{70.9(5)} \\
\cline{1-7}
   \multirow{7}{*}{{\begin{tabular}[c]{@{}c@{}}ViT+BeiT:\\ BerT+DistilBerT\\ (4+4:4+4)\end{tabular}}}& - & Standalone   & - & -&  58.3(2)  &60.2(2) \\
   
\cdashline{2-7}
   
            &\multirow{4}{*}{Logit}   
                         
     & FedET   
 & 74.6(8)  &0.3027(6)&  75.6(6) & 62.3(6)\\
                     &  & FedMD   
 & 73.4(4)  & 0.3015(2) & 73.6(5)  & 61.1(5)\\
                   & & FedGems &75.2(5)&0.3032(3)
 &74.7(4) & 62.0(4) \\
                    &  & l-FedMMKT   &   75.9(7)& 0.3040(3)
&   76.5(6) & 63.0(6)
  \\
  \cdashline{2-7}
 & \multirow{3}{*}{Rep}   
                         & FedMEKT &  73.4(7) &  0.3016(4) & 74.3(5)  &61.9(4)\\
                     &  & CreamFL & 75.5(9) & 0.3036(7)& 76.0(8)& 62.8(7)\\
                    &  & FedMMKT  & \textbf{76.9(6)}& \textbf{0.3055(5)}& \textbf{77.5(7)} & \textbf{63.8(6)}
\\
\bottomrule
\end{tabular}
}
\begin{tablenotes}
\footnotesize
\item[*] All unimodal baselines are adapted to the multimodal setting by separately training on each modality and aggregating results across all clients.
\end{tablenotes}
\label{table3}
\end{table*}
Due to the lack of prior work that fine-tunes a server T2I model in FL setting, comparing both server and client performance in a unified framework is challenging. To address this, we further integrate FedMMKT's T2I model and fine-tuning strategy with several prevalent FL methods, each forming a distinct knowledge transfer loop between a server T2I model and clients. This allows for a direct comparison of their knowledge transfer efficiency. The results are shown in Table~\ref{table3}. First of all, integrating the same server‑side generator improves all methods by comparing Table~\ref{table1} and Table~\ref{table3}, but the magnitude of gain varies depending on efficacy of knowledge transfer. FedMMKT consistently outperforms all baselines on both server-side T2I accuracy and client-side accuracy, thanks to its contrastive, representation-level aggregation that preserves fine-grained multimodal semantics, as well as its LabVote-based reweighting that suppresses noisy clients to enhance server-side data quality. Compared to other logit-based methods such as FedET and FedGems, l-FedMMKT demonstrates stronger performance by more effectively aligning semantic information across heterogeneous multimodal clients. In addition, unimodal baselines (e.g., FedMD, FedMEKT) generally lag behind their multimodal counterparts due to limited cross-modal knowledge exchange. Among multimodal methods, naive feature averaging (e.g., CreamFL, FedMEKT) is not sufficient to resolve domain conflicts, highlighting representation aggregation as a key component for effective multimodal FL. The results of using the Stable Diffusion model as the pre-trained T2I model on the server are shown in Table~\ref{table7}. Table~\ref{table4} presents the comparison results of FedMMKT and baselines using GLIDE on the Food101 dataset.

\begin{table*}[ht]
\centering
\caption{Comparison of FedMMKT with baselines using Stable Diffusion model on Flowers102 dataset}
\resizebox{0.65\linewidth}{!}
{
\begin{tabular}{cccccccccc}
\hline
 \textbf{Client Model} & \textbf{Transfer} & \textbf{Method} & \textbf{T2I Acc} &\textbf{Clip score}  & \textbf{Img Acc} & \textbf{Txt Acc} \\
\hline
 - & - & None & 0.2998(4) & 41.8(9) & - & -  \\
 - & - & Real data &0.3051(4)  & 59.2(9) & - & - \\ 
\hline
\multirow{8}{*}{{\begin{tabular}[c]{@{}c@{}}ViT+BeiT:\\ BerT+DistilBerT\\(2+2:2+2)\end{tabular}}}  &- & Standalone   & - & -&  65.4(1)  &  66.8(3) \\
\cline{2-9}
&\multirow{4}{*}{Logit} 
                         & FedGems &51.6(3) &\textbf{0.3032(4)} &73.8(5)  &68.8(4)  \\
              &  & FedET   & 51.9(5)  &0.3028(3)       & 74.0(6)       &68.9(5)  \\
                &  & FedMD   &50.0(3)  &0.3023(2)      &73.2(4)             &68.4(3) \\
                &  & l-FedMMKT    &52.2(6)  &0.3030(3)            &74.4(5)             &69.2(7)  \\ 
\cline{2-9}
                & \multirow{3}{*}{Rep}   
                         & FedMEKT &50.7(2)  &0.3024(2) &73.4(3)  &68.6(3)  \\
                 &  & CreamFL   &51.3(7)&0.3026(4)          &73.8(6)          &68.9(5)  \\
                 &  & FedMMKT  &\textbf{53.0(8)}  &\textbf{0.3032(4) }       &\textbf{75.2(6)}              &\textbf{69.6(5) } \\  

\hline
\end{tabular}
}
\label{table7}
\end{table*}

\begin{table*}[ht]
\centering
\caption{Comparison of FedMMKT with baselines using GLIDE on Food101 dataset.}
\resizebox{0.65\linewidth}{!}
{
\begin{tabular}{cccccccccc}
\toprule
 \textbf{Client Model} & \textbf{Transfer} & \textbf{Method} & \textbf{T2I Acc} & \textbf{Clip score} & \textbf{Img Acc}  & \textbf{Txt Acc} \\
\cline{1-9}
 \multirow{3}{*}{ViT+BeiT(2+2)}& -  & Standalone   & - & -&  66.2(2) & - \\
  
\cdashline{2-9}
                &\multirow{4}{*}{Logit}  & FedGems  &71.6(8)&0.2955(5)& 78.0(6)& - \\
                 &  & FedET   &68.8(9)&0.2931(7)&   76.4(7) & - \\
                  &  & FedMD  & 69.4(8)&0.2937(5) & 77.7(5) & - & - \\
                  &  & l-FedMMKT  & 72.0(7) & 0.2958(4)  & 78.8(5) & - \\
                   \cdashline{2-9}
  &\multirow{2}{*}{Rep} & FedMEKT & 71.3(6) & 0.2953(6) & 78.0(7) & - \\
                  &  & u-FedMMKT  & \textbf{72.8(4)} & \textbf{0.2966(4)}& \textbf{79.8(5)} & - \\
\cline{1-9}
\multirow{6}{*}{{\begin{tabular}[c]{@{}c@{}}ViT+BeiT:\\ BerT+DistilBerT\\(2+2:2+2)\end{tabular}}}& -  & Standalone   & - & -&  66.2(2) & 65.9(3)\\
   
\cdashline{2-9}
                 &\multirow{4}{*}{Logit} 
                         & FedGems  & 74.3(6) & 0.2983(5)& 80.5(6)& 68.6(5) \\
                 &  & FedET  & 72.6(8)& 0.2964(7)  & 78.7(7) & 67.4(6)\\
                  &  & FedMD  & 73.6(4)& 0.2975(3) & 79.8(5) &   68.0(6) \\
                 &  & l-FedMMKT   & 75.1(5) & 0.2985(3)& 80.6(4) &  69.7(5)\\
                 \cdashline{2-9}
   & \multirow{3}{*}{Rep}   
                         & FedMEKT  & 71.8(6) & 0.2956(5) & 79.9(5)  & 68.2(4) \\
                 &  & creamFL  & 73.2(9) & 0.2973(8)& 80.7(8) & 69.1(7) \\
                 &  & FedMMKT  & \textbf{77.2(6)} & \textbf{0.3004(4) } &  \textbf{81.4(6)} & \textbf{70.0(5)} \\        
\cline{1-9}
\multirow{6}{*}{{\begin{tabular}[c]{@{}c@{}}ViT+BeiT:\\ BerT+DistilBerT\\ (4+4:4+4)\end{tabular}}}& -  & Standalone   & - & -&  61.7(2) & 59.3(4)\\
  
\cdashline{2-9}
         &\multirow{4}{*}{Logit}& FedGems &71.5(7)&0.2954(5) &67.5(6)  & 61.2(5) \\
                  &  & FedET   & 71.2(9) &0.2950(8) &  68.4(7)  & 61.8(7)\\
                  &  & FedMD   & 70.3(4)& 0.2941(4) &   67.0(6)  & 61.2(4)\\
                  &  &l- FedMMKT    &   72.1(5)& 0.2959(4)&  69.5(4)   &62.5(5) \\
                   \cdashline{2-9}
& \multirow{3}{*}{Rep}   
                         & FedMEKT  &  70.2(4) &  0.2934(5)  & 66.2(6)  &61.3(5)\\
                  &  & creamFL  & 70.3(9) & 0.2941(8)& 66.8(9)& 61.5(6)\\
                  &  & FedMMKT  & \textbf{73.3(5) } &\textbf{ 0.2972(6)}&  \textbf{69.9(6)} & \textbf{62.8(4)}\\
\bottomrule
\end{tabular}
}
\label{table4}
\end{table*}

\subsection{Ablation Studies}

\begin{table}[ht]
\centering
\caption{Effectiveness of components in FedMMKT using Flowers102 dataset}
\resizebox{1.0\linewidth}{!}{
\begin{tabular}{ccccccc}
\toprule
 \textbf{Method}  & \textbf{T2I Acc} & \textbf{Clip score}& \textbf{Img Acc} & \textbf{Txt Acc} \\
\midrule
FedMMKT & \textbf{81.2(7)} & \textbf{0.3083(2) }& \textbf{84.1(5)} & \textbf{70.9(5)} \\
-- LabVote  & 73.6(9) & 0.3021(7) & 80.5(7) & 67.7(6) \\
-- MultiRepFusion & 77.0(6)  & 0.3055(5) & 81.6(4) & 70.1(4) \\
-- (LabVote + MultiRepFusion) & 72.5(7) & 0.3010(4) & 78.9(5) & 67.5(5) \\
-- T2I\_Rep  & 78.6(9) & 0.3061(8)& 83.0(9) & 69.7(8)\\
-- Client\_Rep  & 77.4(8) & 0.3048(7)& 81.8(7) & 69.6(8) \\
\bottomrule
\end{tabular}
}
\label{table5}
\end{table}

Table~\ref{table5} presents an ablation study evaluating the contributions of key components. Removing either LabVote or MultiRepFusion leads to notable drops in both T2I and client model accuracy, confirming their importance for effective multimodal alignment and aggregation. We further assess the role of aggregated representation training by disabling it on either the T2I model (“-T2I\_Rep”) or the client models (“-Client\_Rep”). In “-T2I\_Rep”, the server omits aggregated features while clients still use the full objective (Eq.\ref{eq13}); in “-Client\_Rep”, clients are trained only on synthetic data without the KL alignment term from Eq.\ref{eq13}. Results show that both components are critical, highlighting the importance of representation-level knowledge transfer for both sides.
\subsection{Comparison with Parameter-Efficient Fine-Tuning Baselines}

\begin{table}[ht]
\centering
\caption{Comparison of FedMMKT with parameter-efficient fine-tuning baselines}
\resizebox{1.0\linewidth}{!}
{
\begin{tabular}{cccccc}
\hline
 \textbf{Transfer} & \textbf{Method} & \textbf{T2I Acc} & \textbf{Clip score} & \textbf{Img Acc} & \textbf{Txt Acc} \\
\hline
LoRA   & LoRA-FL   &- &- & 57.5(5)  &40.9(7) \\       
Bias   & Bias-FL   &- &- &38.5(9)   &35.7(8) \\
Adapter& FedDAT    &70.9(6)   & 0.2994(2)&69.7(4) & 68.9(4) \\
Logit  & l-FedMMKT  & 77.0(6)& 0.3055(3)& 81.6(4)& 70.1(4) \\
Rep    &  FedMMKT   & \textbf{81.2(7)} & 0.3083(3)& \textbf{84.1(5)}& \textbf{70.9(5)}\\

\hline
\end{tabular}
}
\label{table8}
\end{table}

\begin{table}[ht]
\centering
\caption{Generalization capability of T2I Model}
\label{table6}
\resizebox{0.85\linewidth}{!}{
\small
\begin{tabular}{cccc}
\toprule
 \textbf{Method} & \textbf{FID $\downarrow$} & \textbf{IS $\uparrow$} & \textbf{Clip score $\uparrow$}\\
\midrule
Original GLIDE & 16.77(5) & 20.54(2) & 0.2696(5)\\
FedMMKT-Flowers102
 & 16.82(8) & 20.27(4)  & 0.2687(2)\\
FedMMKT-Food101
 & 16.83(7) & 20.22(7) & 0.2685(3)\\
\bottomrule
\end{tabular}
}
\end{table}

\begin{table}[ht]
\centering
\caption{Comparison of FedMMKT with different synthetic data volume}
\resizebox{1.0\linewidth}{!}{%
\small
\begin{tabular}{ccccc}
\hline
\textbf{Number of synthetic data} &
\textbf{T2I Acc} &
\textbf{Clip score} &
\textbf{Img Acc} &
\textbf{Txt Acc} \\
\hline
0   & $68.3(9)$  & $0.2975(5)$ & $65.4(1)$ & $66.8(3)$ \\
4k  & $73.5(6)$  & $0.3016(3)$ & $74.9(5)$ & $68.6(6)$ \\
6k  & $76.5(5)$  & $0.3047(4)$ & $79.3(3)$ & $69.9(4)$ \\
8k  & $77.0(4)$  & $0.3055(3)$ & $81.6(2)$ & $70.7(2)$ \\
10k & $77.0(4)$  & $0.3055(2)$ & $81.6(2)$ & $70.7(2)$ \\
\hline
\end{tabular}%
}
\label{table9}
\end{table}

\begin{table}[ht]
\centering
\caption{\small The upload and download overhead per iteration under ViT+BeiT: BerT+DistilBerT (2+2:2+2) client setting. }
\resizebox{0.9\linewidth}{!}{
\begin{tabular}{ccccc}
\toprule
\textbf{Method} & \textbf{Upload}& \textbf{Download} & \textbf{T2I Acc}&\textbf{Client Acc}\\
\midrule
FedDAT & 3.0MB & 3.0MB & 70.9(6) & 69.3\\
CreamFL  &  23.4MB &  46.9MB & -& 73.7\\
FedMEKT  &  23.4MB &  23.4MB & -& 74.0\\
l-FedMMKT & 0.31MB &  4.78MB &77.0(6)&75.9 \\
FedMMKT & 2.35MB & 7.13MB &\textbf{81.2(7)} &\textbf{77.5}\\
\bottomrule
\end{tabular}
}
\label{table:comm_cost}
\end{table}
Table~\ref{table8} presents a comparison between FedMMKT and several parameter-efficient fine-tuning baselines, including LoRA-FL, Bias-FL, and FedDAT \cite{50_FedDAT}. The methods "LoRA-FL" and "Bias-FL" are adapted versions of the centralized LoRA and bias-tuning methods, extended into the FL setting by following the baselines used in FedDAT \cite{50_FedDAT}. For FedDAT, we integrated a T2I model on the server side. Inspired by the T2I\_Adapter \cite{53_t2i}, the globally trained adapter from FedDAT was inserted into the UNet decoder of the T2I model. In all these methods, clients were restricted to unimodal data (either image or text), and the server aggregated the full modal adapters or biases using the FedAvg algorithm.

The poor performance of LoRA-FL and Bias-FL can be attributed to lack of design mechanisms to align or merge knowledge from clients with separated modalities.  Although FedDAT outperformed LoRA-FL and Bias-FL by employing mutual distillation designed for heterogeneous tasks, it failed to address the challenge of modality separation effectively. Mutual distillation facilitates knowledge exchange within the same modality but does not bridge the gap between separated modalities. Additionally, FedDAT does not leverage synthetic datasets, which further limits its model performance. These observations underscore the advantages of leveraging synthetic data and the multimodal integration strategy employed in our FedMMKT framework, resulting in enhanced performance when compared to parameter-efficient baselines.

\subsection{Generalization Capability of T2I Model}
Next, we evaluate the generalization ability of the fine-tuned T2I model. Following previous work \cite{1,2}, we test the model on the generic dataset MS-COCO \cite{47} and obtained metrics including FID \cite{48}, Inception Score (IS) \cite{49}, and CLIP score. Table \ref{table6} shows that after fine-tuning using FedMMKT, the FID, IS, and CLIP scores of the model are very close to those of the original T2I model, suggesting minimal degradation in generative quality. This result highlights that FedMMKT enables effective domain adaptation while preserving the model's broader generative abilities.

\subsection{Effect of Synthetic Data Volume on Performance }

We evaluate the impact of synthetic data volume on model performance using the Flowers102 dataset and the GLIDE generative model. As shown in Table~\ref{table9}, increasing the number of synthetic samples from 0k to 8k leads to significant improvements in both image and text accuracy, with image accuracy rising from 65.4\% to 81.6\% and text accuracy increasing from 66.8\% to 70.7\%. The Clip Score and T2I accuracy also improve accordingly, indicating better multimodal alignment.
However, beyond 8k synthetic samples, the performance gain saturates, as seen in the minimal difference between 8k and 10k samples. This suggests that while synthetic data enhances learning, excessive amounts may not provide additional benefits.

\subsection{Communication Cost}
As shown in Table~\ref{table:comm_cost}, FedMMKT achieves significantly lower upload and download costs compared to FedMEKT and CreamFL. This is because  FedMEKT and CreamFL require exchanging representations of the entire public dataset in every round. In contrast, FedMMKT generates a small set of synthetic samples per iteration and transmits only their logits or representations. This design results in the lowest upload cost among all methods, which is particularly advantageous in real-world settings where uplink bandwidth is limited.

\section{Conclusion}
\label{Conclusion}
In this paper, we present FedMMKT, an innovative FL framework for multimodal knowledge transfer that enhances both T2I and client models collaboratively, without sharing client data or model parameters. FedMMKT leverage synthetic data as an intermediary for knowledge transfer, incorporating novel mechanisms to effectively refine data quality and integrate multimodal representations from heterogeneous clients, facilitating bidirectional knowledge transfer between the server-side T2I model and client models. Our work illuminates the benefits and challenges of cross-silo knowledge transfer between resource-constrained data parties and large pre-trained models.

\section*{Acknowledgments}
The authors gratefully acknowledge the financial support provided by National Key R\&D Program of China under Grant (2024YFB4207203), National Natural Science Foundation of China under grant (52401376), Pioneer and Leading Goose R\&D Program of Zhejiang (2025C04005), Quzhou Science and Technology Plan Project (2024K154).
{\appendices
\section{Examples of Label Description Prompt and Textual Description}
\label{Appendix B}
\begin{table*}[h!]
    \centering
    \caption{Examples of Label Description Prompt and Textual Description from the Flowers102 dataset.}
    \label{table6}
    \begin{tabular}{ >{\centering\arraybackslash}m{3cm}| >
    { \centering\arraybackslash}m{6cm}| >{\centering\arraybackslash}m {6cm}}
        \hline
        \textbf{Class Label($\mathcal{Y}^s$)} & \textbf{Label Description Prompt($P(\mathcal{Y}^s)$}) & \textbf{Textual Description ($T^s$)} \\ \hline
        Canterbury Bells & A picture of Canterbury bells, also named Campanula medium.
        & The flower is large and pink, bell-shaped with a prominent white stamen at its center. The surrounding leaves are green, slender, and bell-shaped, providing a graceful frame to the flower. 
        \\ \hline
        English Marigold & A picture of English marigold, also named Calendula officinalis. 
        & The flower is large and bright orange, with a distinct dark center. It is surrounded by dark green leaves, providing a rich contrast and enhancing the vibrancy of the orange petals. 
        \\ \hline
        Tiger Lily & A picture of a Tiger lily.
        & The flower with large, vivid orange petals marked by distinctive dark spots. The petals gracefully droop from long, slender stems, characteristic of the lily family. \\ \hline
        Moon Orchid & A picture of a Moon orchid.
        & The flower is characterized by its large, round, white flowers, occasionally tinged with light pink at the center. The plant also features broad, glossy leaves and a single slender stem supporting the flowers. \\ \hline
        Bird of Paradise & A picture of a Bird of paradise.
        & The flower is distinguished by its striking orange and blue petals, resembling a bird's head. The flower emerges from a boat-shaped bract, complemented by long, banana-like green leaves typical of tropical regions. \\ \hline
    \end{tabular}    
\end{table*}

Table~\ref{table6} provides several examples of label description prompt ($P(\mathcal{Y}^s)$) and textual description ($T^s$) from the Flowers102 dataset. Here, $\mathcal{Y}^s$ represents the class label, while $P(\mathcal{Y}^s)$ is used to guide the T2I model to generate images by providing a concise and clear specification of the desired image content. In contrast, $T^s$ is generated by the Blip model after the image generation, offering a detailed description to enrich the semantic information of the image.

\section{Comparison of LabVote Strategies on the Effectiveness of Label Correction for T2I Models}
\label{Appendix C}
\begin{figure*}[ht]
  \centering
  \includegraphics[width=1.1\linewidth]{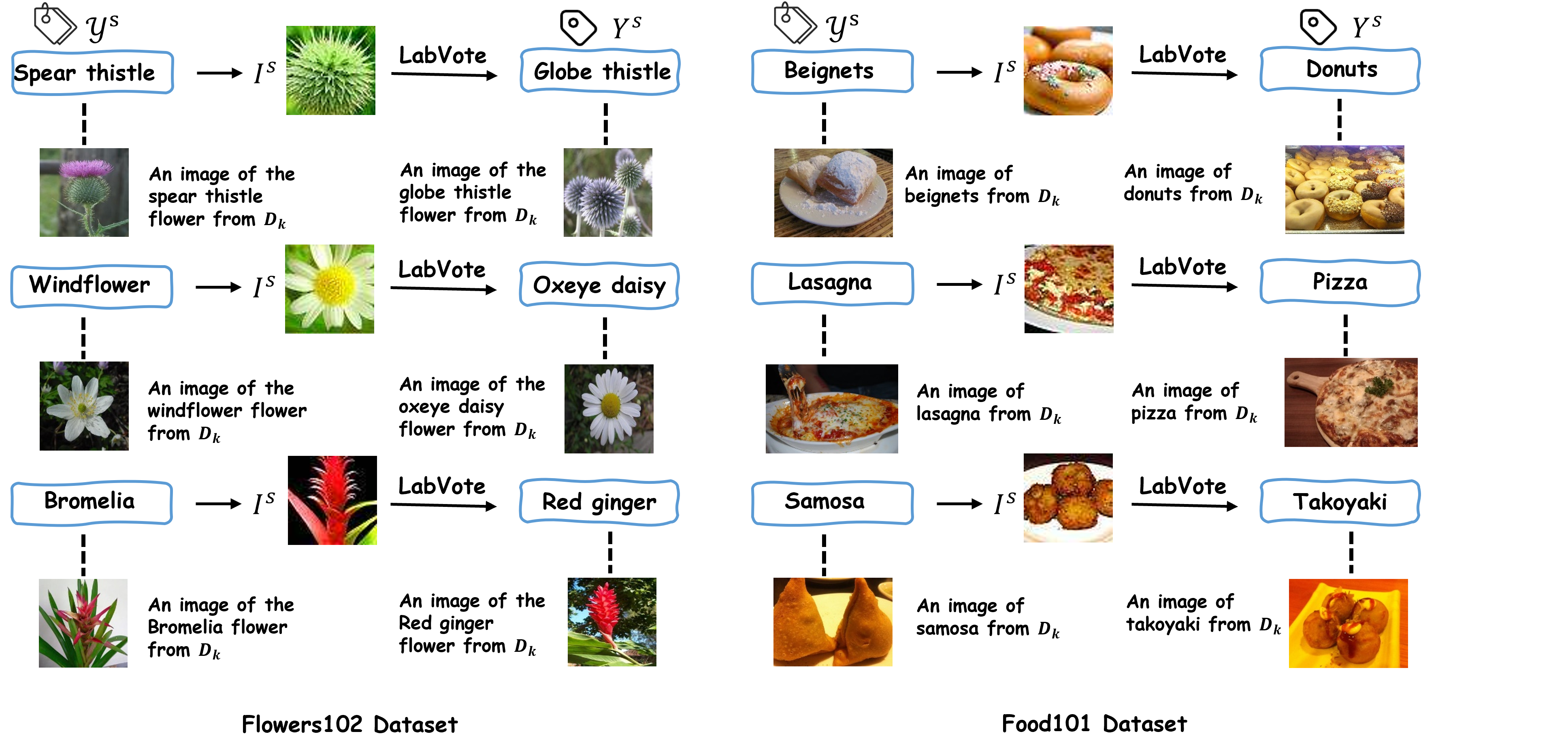}
  \caption{Some of the images generated by the T2I model with label bias and LabVote corrected label visualisation}
  \label{Fig.4}
\end{figure*}

Fig.~\ref{Fig.4} illustrates the label bias present in images generated by T2I models. The $\mathcal{Y}^s$ represent the class labels, while $I^s$ are the images generated by the T2I model based on these labels. Upon inspection, we observe that the visual characteristics of $I^s$ do not align with those of the images associated with $\mathcal{Y}^s$, which were randomly selected from the client dataset $D_k$. From this visual comparison, the T2I model may generate images that do not match their intended labels within specific domains. After applying the LabVote algorithm, the corrected labels denoted as ${Y}^s$. We find that the visual characteristics of $I^s$ are much closer to those of the client dataset's images corresponding to ${Y}^s$, indicating that the LabVote algorithm effectively mitigates the label bias.

\section{Logit-based FedMMKT} 
\label{Appendix D}
The logit-based variant of FedMMKT transmits logits instead of representations between clients and server. The Cross-Client Contrastive Aggregation reduces to: 
\small
\begin{equation}
w_i^{(n)} = \log \left( \frac{\exp(\scriptstyle{\text{Cos}(\text{logit}_i^{(n)}, \frac{1}{K-N} \sum_{t=1}^{K-N} \text{logit}_i^{(t)})})}{\sum_{j=1}^{\text{batch}} 1_{[j \neq i]} \exp(\scriptstyle{\text{Cos}(\text{logit}_i^{(n)}, \frac{1}{K-N} \sum_{t=1}^{K-N} \text{logit}_j^{(t)})})} \right)
\label{14}
\end{equation}
\normalsize

The server then aggregates the logits using:
\small
\begin{equation}
\alpha_i^{(1)}, \alpha_i^{(2)}, \dots, \alpha_i^{(K)} = \text{softmax}(w_i^{(1)}, w_i^{(2)}, \dots, w_i^{(K)})
\label{15}
\end{equation}
\normalsize

\begin{equation}
\text{logit}_i^{s} = \sum_{k=1}^K \alpha_i^{(k)} \cdot \text{logit}_i^{(k)}
\label{16}
\end{equation}

Once the logits are aggregated, the consensus label \( Y_i^{s} \) can be calculated based on \( \text{logit}_i^{s} \):

\begin{equation}
Y_i^s = \arg\max_{t \in \mathcal{T}} (\text{logit}_i^s[t])
\label{17}
\end{equation}

The server-side T2I model then directly leverages the synthetic data selected using the aggregated logits. The T2I model is fine-tuned with these high-quality synthetic samples using only textual prompts and consensus labels, focusing on optimizing the decoder \( d(\cdot) \).

The client models are then trained with the following cross-entropy loss:
\begin{equation}
    L_{local} = -\sum_{i=1}^{|D^{s'}|} l(\mathcal{M'}_k(x_i^s),Y_i^{s})\label{18}
\end{equation}

The logit-based variant bypasses the detailed representation alignment by focusing directly on the distilled prediction outputs from client models, reducing computational and communication costs. This offers an efficient, lightweight solution for server-client collaboration.

\section{Unimodal FedMMKT}
FedMMKT can also reduce to an unimodal variant. In scenarios where all clients possess data of the same modality, FedMMKT bypasses the Inter-Modal Cross-Attention step, and performs Cross-Client Contrastive Aggregation as the following: 

\small
\begin{equation}
w_i^{(n)} = \log \left( \frac{\exp(\scriptstyle{\operatorname{Cos}(\mathbf{RI}_i^{(n)}, \mathbf{RI}_i^{(n,z)})})}{\sum_{j=1}^{\text{batch}} \mathbf{1}_{[j \neq i]} \exp(\scriptstyle{\operatorname{Cos} (\mathbf{RI}_i^{(n)}, \mathbf{RI}_j^{(n,z)})})} \right) 
\label{19}
\end{equation}
\normalsize

where, $\mathbf{RI}^{(n)}$ denote original representations, $\mathbf{RI}^{(n,z)}$ denote enhanced representations generated from the original data by applying transformations such as random cropping and color distortion, inspired by SimCLR\cite{28}. The same LabVote process as in multimodal scenarios is employed. 

\section{Per-Round Communication Cost}
\label{Comm_Calc}

We compute per-round upload and download costs (in MB) under the following configuration:

\begin{itemize}
  \item Number of clients: $K=8$, with $N=4$ image clients and $K{-}N=4$ text clients.
  \item Float size: $B = 4$ bytes.
  \item Synthetic samples: $|D^s| = 100$.
  \item Class count: $C = 102$.
  \item Representation dimension: $d = 768$.
  \item Synthetic image size: $S_I = 64 \times 64 \times 3 = 12{,}288$ bytes.
  \item Synthetic text size (token IDs): $S_T = 64 \times 4 = 256$ bytes.
\end{itemize}

The upload and download costs are computed as follows:
\paragraph{1-FedMMKT:}
\[\text{Upload} = |D^s| \cdot K \cdot B \cdot C \]
\[\text{Download} = |D^s| \cdot S_I \cdot N + |D^s| \cdot S_T \cdot (K{-}N) + |D^s| \cdot K \cdot B\]

\paragraph{FedMMKT:}
\[\text{Upload} = |D^s| \cdot K \cdot B \cdot (d + 2)\]
\[\text{Download} = |D^s| \cdot S_I \cdot N + |D^s| \cdot S_T \cdot (K{-}N) + |D^s| \cdot K \cdot B \cdot (d + 1)\]

}

\vfill
\typeout{get arXiv to do 4 passes: Label(s) may have changed. Rerun}

\begin{thebibliography}{1}
\bibliographystyle{IEEEtran}
\bibitem{1}
A. Nichol, P. Dhariwal, A. Ramesh, P. Shyam, P. Mishkin, B. McGrew, I. Sutskever, and M. Chen, 
``{\it Glide: Towards photorealistic image generation and editing with text-guided diffusion models},'' 
{\it arXiv preprint arXiv:2112.10741}, 2021.

\bibitem{2}
A. Ramesh, P. Dhariwal, A. Nichol, C. Chu, and M. Chen, 
``{\it Hierarchical text-conditional image generation with CLIP latents},'' 
{\it arXiv preprint arXiv:2204.06125}, 2022.

\bibitem{3}
R. Rombach, A. Blattmann, D. Lorenz, P. Esser, and B. Ommer, 
``{\it High-resolution image synthesis with latent diffusion models},'' 
in {\it Proc. IEEE/CVF Conf. Comput. Vis. Pattern Recognit. (CVPR)}, 
pp. 10684--10695, 2022.

\bibitem{4}
A. Rahman, M. S. Hossain, G. Muhammad, D. Kundu, T. Debnath, M. Rahman, M. S. I. Khan, P. Tiwari, and S. S. Band, 
``{\it Federated learning-based AI approaches in smart healthcare: concepts, taxonomies, challenges and open issues},'' 
{\it Cluster Comput.}, vol. 26, no. 4, pp. 2271--2311, 2023.

\bibitem{5}
Q. Yang, Y. Liu, T. Chen, and Y. Tong, 
``{\it Federated machine learning: Concept and applications},'' 
{\it ACM Trans. Intell. Syst. Technol. (TIST)}, vol. 10, no. 2, pp. 1--19, 2019.

\bibitem{6}
T. Li, A. K. Sahu, A. Talwalkar, and V. Smith, 
``{\it Federated learning: Challenges, methods, and future directions},'' 
{\it IEEE Signal Process. Mag.}, vol. 37, no. 3, pp. 50--60, 2020.

\bibitem{7}
H. Zhao, W. Du, F. Li, P. Li, and G. Liu, 
``{\it Reduce communication costs and preserve privacy: Prompt tuning method in federated learning},'' 
{\it arXiv preprint arXiv:2208.12268}, 2022.

\bibitem{8}
Z. Zhang, Y. Yang, Y. Dai, L. Qu, and Z. Xu, 
``{\it When Federated Learning Meets Pre-trained Language Models' Parameter-Efficient Tuning Methods},'' 
{\it arXiv preprint arXiv:2212.10025}, 2022.

\bibitem{9}
S. Babakniya, A. R. Elkordy, Y. H. Ezzeldin, Q. Liu, K.-B. Song, M. El-Khamy, and S. Avestimehr, 
``{\it SLoRA: Federated parameter efficient fine-tuning of language models},'' 
{\it arXiv preprint arXiv:2308.06522}, 2023.

\bibitem{10}
Y. Tian, Y. Wan, L. Lyu, D. Yao, H. Jin, and L. Sun, 
``{\it FedBERT: When federated learning meets pre-training},'' 
{\it ACM Trans. Intell. Syst. Technol. (TIST)}, vol. 13, no. 4, pp. 1--26, 2022.

\bibitem{11}
Y. Deng, Z. Qiao, J. Ren, Y. Liu, and Y. Zhang, 
``{\it Mutual enhancement of large and small language models with cross-silo knowledge transfer},'' 
{\it arXiv preprint arXiv:2312.05842}, 2023.

\bibitem{12_vit}
A. Dosovitskiy, 
``{\it An image is worth 16x16 words: Transformers for image recognition at scale},'' 
{\it arXiv preprint arXiv:2010.11929}, 2020.

\bibitem{13_beit}
H. Bao, L. Dong, S. Piao, and F. Wei, 
``{\it Beit: Bert pre-training of image transformers},'' 
{\it arXiv preprint arXiv:2106.08254}, 2021.

\bibitem{14_bert}
J. Devlin, 
``{\it Bert: Pre-training of deep bidirectional transformers for language understanding},'' 
{\it arXiv preprint arXiv:1810.04805}, 2018.

\bibitem{15_distilbert}
V. Sanh, 
``{\it DistilBERT, A Distilled Version of BERT: Smaller, Faster, Cheaper and Lighter},'' 
{\it arXiv preprint arXiv:1910.01108}, 2019.


\bibitem{16_flowers102}
M.-E. Nilsback and A. Zisserman, 
``{\it Automated flower classification over a large number of classes},'' 
in {\it 2008 Sixth Indian Conference on Computer Vision, Graphics \& Image Processing}, 
IEEE, pp. 722--729, 2008.

\bibitem{17_food101}
L. Bossard, M. Guillaumin, and L. Van Gool, 
``{\it Food-101--mining discriminative components with random forests},'' 
in {\it Computer Vision--ECCV 2014: 13th European Conference, Zurich, Switzerland, September 6-12, 2014, Proceedings, Part VI 13}, 
Springer, pp. 446--461, 2014.

\bibitem{18}
B. Xiong, X. Yang, F. Qi, and C. Xu, 
``{\it A unified framework for multi-modal federated learning},'' 
{\it Neurocomputing}, vol. 480, pp. 110--118, 2022.

\bibitem{19_fedsea}
M. Tan, Y. Feng, L. Chu, J. Shi, R. Xiao, H. Tang, and J. Yu, 
``{\it FedSea: Federated Learning via Selective Feature Alignment for Non-IID Multimodal Data},'' 
{\it IEEE Transactions on Multimedia}, 2023.

\bibitem{20_fedmfs}
L. Yuan, D.-J. Han, V. P. Chellapandi, S. H. Zak, and C. G. Brinton, 
``{\it Fedmfs: Federated multimodal fusion learning with selective modality communication},'' 
in {\it ICC 2024-IEEE International Conference on Communications}, 
IEEE, pp. 287--292, 2024.
\bibitem{21_creamFL}
Q. Yu, Y. Liu, Y. Wang, K. Xu, and J. Liu, 
``{\it Multimodal federated learning via contrastive representation ensemble},'' 
{\it arXiv preprint arXiv:2302.08888}, 2023.

\bibitem{22_fedmekt}
H. Q. Le, M. NH. Nguyen, C. M. Thwal, Y. Qiao, C. Zhang, and C. S. Hong, 
``{\it Fedmekt: Distillation-based embedding knowledge transfer for multimodal federated learning},'' 
{\it arXiv preprint arXiv:2307.13214}, 2023.

\bibitem{23}
W. Feng, X. He, T.-J. Fu, V. Jampani, A. Akula, P. Narayana, S. Basu, X. E. Wang, and W. Y. Wang, 
``{\it Training-free structured diffusion guidance for compositional text-to-image synthesis},'' 
{\it arXiv preprint arXiv:2212.05032}, 2022.

\bibitem{24_blip}
J. Li, D. Li, C. Xiong, and S. Hoi, 
``{\it Blip: Bootstrapping language-image pre-training for unified vision-language understanding and generation},'' 
in {\it International Conference on Machine Learning}, 
PMLR, pp. 12888--12900, 2022.

\bibitem{25}
L. Che, J. Wang, Y. Zhou, and F. Ma, 
``{\it Multimodal federated learning: A survey},'' 
{\it Sensors}, vol. 23, no. 15, pp. 6986, 2023.

\bibitem{26}
Y.-M. Lin, Y. Gao, M.-G. Gong, S.-J. Zhang, Y.-Q. Zhang, and Z.-Y. Li, 
``{\it Federated learning on multimodal data: A comprehensive survey},'' 
{\it Machine Intelligence Research}, vol. 20, no. 4, pp. 539--553, 2023.

\bibitem{27}
T. Feng, D. Bose, T. Zhang, R. Hebbar, A. Ramakrishna, R. Gupta, M. Zhang, S. Avestimehr, and S. Narayanan, 
``{\it Fedmultimodal: A benchmark for multimodal federated learning},'' 
in {\it Proceedings of the 29th ACM SIGKDD Conference on Knowledge Discovery and Data Mining}, 
pp. 4035--4045, 2023.

\bibitem{28}
T. Chen, S. Kornblith, M. Norouzi, and G. Hinton, 
``{\it A simple framework for contrastive learning of visual representations},'' 
in {\it International Conference on Machine Learning}, 
PMLR, pp. 1597--1607, 2020.

\bibitem{29_dl}
T.-M. H. Hsu, H. Qi, and M. Brown, 
``{\it Measuring the effects of non-identical data distribution for federated visual classification},'' 
{\it arXiv preprint arXiv:1909.06335}, 2019.

\bibitem{30_fedmd}
D. Li and J. Wang, 
``{\it Fedmd: Heterogenous federated learning via model distillation},'' 
{\it arXiv preprint arXiv:1910.03581}, 2019.

\bibitem{31_fedgems}
S. Cheng, J. Wu, Y. Xiao, and Y. Liu, 
``{\it Fedgems: Federated learning of larger server models via selective knowledge fusion},'' 
{\it arXiv preprint arXiv:2110.11027}, 2021.

\bibitem{32_fedet}
Y. J. Cho, A. Manoel, G. Joshi, R. Sim, and D. Dimitriadis, 
``{\it Heterogeneous ensemble knowledge transfer for training large models in federated learning},'' 
{\it arXiv preprint arXiv:2204.12703}, 2022.

\bibitem{33}
J. Zhang, Y. Liu, Y. Hua, and J. Cao, 
``{\it An Upload-Efficient Scheme for Transferring Knowledge From a Server-Side Pre-trained Generator to Clients in Heterogeneous Federated Learning},'' 
in {\it Proceedings of the IEEE/CVF Conference on Computer Vision and Pattern Recognition}, 
pp. 12109--12119, 2024.

\bibitem{34_FedMKT}
T. Fan, G. Ma, Y. Kang, H. Gu, L. Fan, and Q. Yang, 
``{\it FedMKT: Federated Mutual Knowledge Transfer for Large and Small Language Models},'' 
{\it arXiv preprint arXiv:2406.02224}, 2024.

\bibitem{35}
L. Qu, W. Wang, Y. Li, H. Zhang, L. Nie, and T.-S. Chua, 
``{\it Discriminative probing and tuning for text-to-image generation},'' 
in {\it Proceedings of the IEEE/CVF Conference on Computer Vision and Pattern Recognition}, 
pp. 7434--7444, 2024.

\bibitem{36}
T. Wang, T. Zhang, B. Zhang, H. Ouyang, D. Chen, Q. Chen, and F. Wen, 
``{\it Pretraining is all you need for image-to-image translation},'' 
{\it arXiv preprint arXiv:2205.12952}, 2022.

\bibitem{37}
A. Brunete, E. Gambao, M. Hernando, and R. Cedazo, 
``{\it Smart assistive architecture for the integration of IoT devices, robotic systems, and multimodal interfaces in healthcare environments},'' 
{\it Sensors}, vol. 21, no. 6, pp. 2212, 2021.

\bibitem{38}
S. Niu, J. Ma, L. Bai, Z. Wang, L. Guo, and X. Yang, 
``{\it EHR-KnowGen: Knowledge-enhanced multimodal learning for disease diagnosis generation},'' 
{\it Information Fusion}, vol. 102, pp. 102069, 2024.

\bibitem{39}
P. Xu, X. Zhu, and D. A. Clifton, 
``{\it Multimodal learning with transformers: A survey},'' 
{\it IEEE Transactions on Pattern Analysis and Machine Intelligence}, vol. 45, no. 10, pp. 12113--12132, 2023.

\bibitem{40}
Y. Zhang, K. Gong, K. Zhang, H. Li, Y. Qiao, W. Ouyang, and X. Yue, 
``{\it Meta-transformer: A unified framework for multimodal learning},'' 
{\it arXiv preprint arXiv:2307.10802}, 2023.
\bibitem{41}
A. Ramesh, M. Pavlov, G. Goh, S. Gray, C. Voss, A. Radford, M. Chen, and I. Sutskever, 
``{\it Zero-shot text-to-image generation},'' 
in {\it International Conference on Machine Learning}, 
PMLR, pp. 8821--8831, 2021.

\bibitem{42}
K. Lee, H. Liu, M. Ryu, O. Watkins, Y. Du, C. Boutilier, P. Abbeel, M. Ghavamzadeh, and S. S. Gu, 
``{\it Aligning text-to-image models using human feedback},'' 
{\it arXiv preprint arXiv:2302.12192}, 2023.

\bibitem{43_dreamsync}
J. Sun, D. Fu, Y. Hu, S. Wang, R. Rassin, D.-C. Juan, D. Alon, C. Herrmann, S. van Steenkiste, R. Krishna, et al., 
``{\it Dreamsync: Aligning text-to-image generation with image understanding feedback},'' 
in {\it Synthetic Data for Computer Vision Workshop@ CVPR 2024}, 2023.

\bibitem{44}
Y. Wan, A. Subramonian, A. Ovalle, Z. Lin, A. Suvarna, C. Chance, H. Bansal, R. Pattichis, and K.-W. Chang, 
``{\it Survey of Bias In Text-to-Image Generation: Definition, Evaluation, and Mitigation},'' 
{\it arXiv preprint arXiv:2404.01030}, 2024.

\bibitem{45}
K. Shmelkov, C. Schmid, and K. Alahari, 
``{\it How good is my GAN?}'' 
in {\it Proceedings of the European Conference on Computer Vision (ECCV)}, 
pp. 213--229, 2018.

\bibitem{46}
A. Radford, J. W. Kim, C. Hallacy, A. Ramesh, G. Goh, S. Agarwal, G. Sastry, A. Askell, P. Mishkin, J. Clark, et al., 
``{\it Learning transferable visual models from natural language supervision},'' 
in {\it International Conference on Machine Learning}, 
PMLR, pp. 8748--8763, 2021.

\bibitem{47}
T.-Y. Lin, M. Maire, S. Belongie, J. Hays, P. Perona, D. Ramanan, P. Dollár, and C. L. Zitnick, 
``{\it Microsoft coco: Common objects in context},'' 
in {\it Computer Vision--ECCV 2014: 13th European Conference, Zurich, Switzerland, September 6-12, 2014, Proceedings, Part V 13}, 
Springer, pp. 740--755, 2014.

\bibitem{48}
M. Heusel, H. Ramsauer, T. Unterthiner, B. Nessler, and S. Hochreiter, 
``{\it Gans trained by a two time-scale update rule converge to a local nash equilibrium},'' 
{\it Advances in Neural Information Processing Systems}, vol. 30, 2017.

\bibitem{49}
T. Salimans, I. Goodfellow, W. Zaremba, V. Cheung, A. Radford, and X. Chen, 
``{\it Improved techniques for training gans},'' 
{\it Advances in Neural Information Processing Systems}, vol. 29, 2016.

\bibitem{50_FedDAT}
H. Chen, Y. Zhang, D. Krompass, J. Gu, and V. Tresp, 
``{\it Feddat: An approach for foundation model finetuning in multi-modal heterogeneous federated learning},'' 
in {\it Proceedings of the AAAI Conference on Artificial Intelligence}, 
vol. 38, no. 10, pp. 11285--11293, 2024.
\bibitem{51_DomainStudio}
J. Zhu, H. Ma, J. Chen, and J. Yuan, 
``{\it DomainStudio: fine-tuning diffusion models for domain-driven image generation using limited data},'' 
2023.

\bibitem{52_dreambooth}
N. Ruiz, Y. Li, V. Jampani, Y. Pritch, M. Rubinstein, and K. Aberman, 
``{\it Dreambooth: Fine tuning text-to-image diffusion models for subject-driven generation},'' 
in {\it Proceedings of the IEEE/CVF Conference on Computer Vision and Pattern Recognition}, 
pp. 22500--22510, 2023.

\bibitem{53_t2i}
C. Mou, X. Wang, L. Xie, Y. Wu, J. Zhang, Z. Qi, and Y. Shan, 
``{\it T2i-adapter: Learning adapters to dig out more controllable ability for text-to-image diffusion models},'' 
in {\it Proceedings of the AAAI Conference on Artificial Intelligence}, 
vol. 38, no. 5, pp. 4296--4304, 2024.

\bibitem{54_diffusiongpt}
J. Qin, J. Wu, W. Chen, Y. Ren, H. Li, H. Wu, X. Xiao, R. Wang, and S. Wen, 
``{\it Diffusiongpt: LLM-driven text-to-image generation system},'' 
{\it arXiv preprint arXiv:2401.10061}, 2024.

\bibitem{55_offsite}
G. Xiao, J. Lin, and S. Han, 
``{\it Offsite-tuning: Transfer learning without full model},'' 
{\it arXiv preprint arXiv:2302.04870}, 2023.

\bibitem{56_fedsp}
Y. Wang, Y. Song, D. Jiang, Y. Ding, X. Wang, Y. Liu, and Q. Liao, 
``{\it FedSP: Federated Speaker Verification with Personal Privacy Preservation},'' 
in {\it International Conference on Algorithms and Architectures for Parallel Processing}, 
Springer, pp. 462--478, 2021.

\bibitem{57_distilling}
G. Hinton, 
``{\it Distilling the Knowledge in a Neural Network},'' 
{\it arXiv preprint arXiv:1503.02531}, 2015.

\bibitem{58_fedavg}
B. McMahan, E. Moore, D. Ramage, S. Hampson, and B. A. y Arcas, 
``{\it Communication-efficient learning of deep networks from decentralized data},'' 
in {\it Artificial Intelligence and Statistics}, 
PMLR, pp. 1273--1282, 2017.

\bibitem{59_promptfl}
T. Guo, S. Guo, J. Wang, X. Tang, and W. Xu, 
``{\it Promptfl: Let federated participants cooperatively learn prompts instead of models-federated learning in age of foundation model},'' 
{\it IEEE Transactions on Mobile Computing}, 2023.

\bibitem{60_pfedprompt}
T. Guo, S. Guo, and J. Wang, 
``{\it Pfedprompt: Learning personalized prompt for vision-language models in federated learning},'' 
in {\it Proceedings of the ACM Web Conference 2023}, 
pp. 1364--1374, 2023.
\bibitem{61_FlexLoRA}
J. Bai, D. Chen, B. Qian, L. Yao, and Y. Li, 
``{\it Federated fine-tuning of large language models under heterogeneous language tasks and client resources},'' 
{\it arXiv e-prints}, pp. arXiv--2402, 2024.

\bibitem{62_fedmbridge}
J. Chen and A. Zhang, 
``{\it FedMBridge: bridgeable multimodal federated learning},'' 
in {\it Forty-first International Conference on Machine Learning}, 2024.

\bibitem{63_disentanglement}
J. Chen and A. Zhang, 
``{\it On disentanglement of asymmetrical knowledge transfer for modality-task agnostic federated learning},'' 
in {\it Proceedings of the AAAI Conference on Artificial Intelligence}, 
vol. 38, no. 10, pp. 11311--11319, 2024.

\bibitem{64_roberta}
Y. Liu, M. Ott, N. Goyal, J. Du, M. Joshi, D. Chen, O. Levy, M. Lewis, L. Zettlemoyer, and V. Stoyanov, 
``{\it Roberta: A robustly optimized bert pretraining approach},'' 
{\it arXiv preprint arXiv:1907.11692}, 2019.

\bibitem{65_differential}
C. Dwork, 
``{\it Differential privacy},'' 
in {\it International Colloquium on Automata, Languages, and Programming}, 
Springer, pp. 1--12, 2006.

\bibitem{zhuang2024foundationmodelmeetsfederated}
W. Zhuang, C. Chen, and L. Lyu, 
``{\it When Foundation Model Meets Federated Learning: Motivations, Challenges, and Future Directions},'' 
{\it arXiv preprint arXiv:2306.15546}, 2024.

\bibitem{zhang-etal-2023-fedpetuning}
Z. Zhang, Y. Yang, Y. Dai, Q. Wang, Y. Yu, L. Qu, and Z. Xu, 
``{\it FedPETuning: When Federated Learning Meets the Parameter-Efficient Tuning Methods of Pre-trained Language Models},'' 
in {\it Findings of the Association for Computational Linguistics: ACL 2023}, 
Association for Computational Linguistics, pp. 9963--9977, Jul. 2023.

\bibitem{wu2024fedbiotllmlocalfinetuning}
F. Wu, Z. Li, Y. Li, B. Ding, and J. Gao, 
``{\it FedBiOT: LLM Local Fine-tuning in Federated Learning without Full Model},'' 
{\it arXiv preprint arXiv:2406.17706}, 2024.

\bibitem{peng2024fedpftfederatedproxyfinetuning}
Z. Peng, X. Fan, Y. Chen, Z. Wang, S. Pan, C. Wen, R. Zhang, and C. Wang, 
``{\it FedPFT: Federated Proxy Fine-Tuning of Foundation Models},'' 
{\it arXiv preprint arXiv:2404.11536}, 2024.

\bibitem{Zhang_2024_CVPR}
J. Zhang, Y. Liu, Y. Hua, and J. Cao, 
``{\it An Upload-Efficient Scheme for Transferring Knowledge From a Server-Side Pre-trained Generator to Clients in Heterogeneous Federated Learning},'' 
in {\it Proceedings of the IEEE/CVF Conference on Computer Vision and Pattern Recognition (CVPR)}, 
pp. 12109--12119, Jun. 2024.

\bibitem{FedProx} Li T, Sahu A K, Zaheer M, et al. Federated optimization in heterogeneous networks. Proceedings of Machine learning and systems, 2020, 2: 429-450.
\bibitem{FedGKD} Yao D, Pan W, Dai Y, et al. FedGKD: Toward heterogeneous federated learning via global knowledge distillation. IEEE Transactions on Computers, 2023, 73(1): 3-17.
\bibitem{FedSSD} Kwan H M, Song S. FedSSD: Scalable and diversity-enhanced distillation for model aggregation in federated learning. arXiv preprint arXiv:2312.17029, 2023.
\bibitem{LipParam} Honorio J. Lipschitz parametrization of probabilistic graphical models. arXiv preprint arXiv:1202.3733, 2012.
\bibitem{Fedcola} Sun G, Mendieta M, Dutta A, et al. Towards multi-modal transformers in federated learning. In: European Conference on Computer Vision. Cham: Springer Nature Switzerland, 2024: 229--246.

\end{thebibliography}
\end{document}